\documentclass[preprint,12pt]{elsarticle}

\usepackage{amssymb}

\usepackage{cite}
\usepackage{amsmath,amssymb,amsfonts}
\usepackage{algorithmic}
\usepackage{graphicx}
\usepackage{textcomp}
\usepackage{xcolor}

\usepackage{multirow}

\usepackage[ruled,linesnumbered]{algorithm2e}

\usepackage{graphicx}
\usepackage{float}
\usepackage{subfigure}

\usepackage{booktabs}    
\usepackage{stfloats}

\usepackage{array}
\usepackage{booktabs}
\usepackage{setspace}

\usepackage{soul}
\soulregister{\cite}7 %
\soulregister{\citep}7 %
\soulregister{\citet}7 %
\soulregister{\ref}7 %
\soulregister{\pageref}7 %

\usepackage[section]{placeins}

\usepackage[colorlinks=blue,
            linkcolor=blue,
            anchorcolor=blue,
            citecolor=blue]{hyperref}
            
\journal{Pattern Recognition}

\begin{document}
\doublespacing
\begin{frontmatter}

\title{Multi-task Meta Label Correction for Time Series Prediction}

\author[inst1]{Luxuan Yang}
\ead{luxuan\_yang@hust.edu.cn}
\affiliation[inst1]{organization={School of Mathematics and Statistics \& Center for Mathematical Sciences},
            addressline={\\Huazhong University of Science and Technology}, 
            city={Wuhan},
            postcode={430074}, 
            country={China}}

\author[inst1]{Ting Gao\corref{cor1}}
\ead{tgao0716@hust.edu.cn}
\cortext[cor1]{Corresponding author}

\author[inst2]{Wei Wei}
\ead{weiw_sjtu@sjtu.edu.cn}
\affiliation[inst2]{organization={Institute of Natural Sciences},
            addressline={Shanghai Jiao Tong University}, 
            city={Shanghai},
            postcode={200240},
            country={China}}
\author[inst3]{Min Dai}
\ead{mindai@whut.edu.cn}
\affiliation[inst3]{organization={School of Science},
            addressline={Wuhan University of Technology}, 
            city={Wuhan},
            postcode={430070},
            country={China}}

\author[inst1]{Cheng Fang}
\ead{fangcheng1@hust.edu.cn}

\author[inst5,inst6]{Jinqiao Duan}
\ead{duan@gbu.edu.cn}
\affiliation[inst5]{organization={Department of Mathematics and Department of Physics},
            addressline={Great Bay University }, 
            city={Dongguan},
            postcode={523000},
            country={China}}
            
\affiliation[inst6]{organization={Dongguan Key Laboratory for Data Science and Intelligent Medicine},
            city={Dongguan},
            postcode={523000},
            country={China}}
\begin{abstract}

Time series classification faces two unavoidable problems. One is partial feature information and the other is poor label quality, which may affect model performance. To address the above issues, we create a label correction method to time series data with meta-learning under a multi-task framework. There are three main contributions. First, we train the label correction model with a two-branch neural network in the outer loop. While in the model-agnostic inner loop, we use pre-existing classification models in a multi-task way and jointly update the meta-knowledge so as to help us achieve adaptive labeling on complex time series. Second, we devise new data visualization methods for both image patterns of the historical data and data in the prediction horizon. Finally, we test our method with various financial datasets, including XOM, S\&P500, and SZ50. Results show that our method is more effective and accurate than some existing label correction techniques.

\end{abstract}


\end{frontmatter}

\section{Introduction}

Nowadays, deep learning has outperformed many conventional methods in various research fields,  especially long term predication of complex time series data. However, there are still two main challenges, insufficient feature extraction and uncertain label quality.

For the first challenge, some efforts have been made to extract more effective feature information from time series data. For instance, Wang and Oates \citep{WangGAF} establish Gramian Angular Field (GAF) and Markov Transition Field methods to encode time series as images to obtain multi-level features. Moreover, a method for visualizing the behavior of nonlinear dynamical systems called Recurrence Plot (RP) \citep{eckmann1995recurrence} is also applied to time series in real-world applications. To name a few examples, Shankar et al. \citep{shankar2021analysis} apply 2D recurrence plot to electroencephalography (EEG) data and train a convolutional neural network (CNN) to analyze epileptic seizures. Barra et al. \citep{barra2020deep} apply GAF for financial data to train an ensemble of CNNs. Chen et al. \citep{chen2016financial} compare the effectiveness of several visualization methods, such as Candlestick Chart, GAF, Moving Average Mapping, and others. They find that GAF achieves the highest accuracy. Sezer and Ozbayoglu \citep{sezer2018algorithmic} also convert stock time series data into some 2-D images and create a trading model called CNN-TA to offer Buy–Sell–Hold signals. Nevertheless, researchers find that some visualization methods such as RP exist a tendency confusion problem\citep{ZHANG2022108385}. To avoid this issue, we provide some enhanced transformation methods for the feature data without neglecting tendency characteristics. The patterns generated from historical features are called ``X" in the following section. \footnote{Here, X is a unified representation that can be $\mathbf{X}^{\prime}$ in noisy data and $\mathbf{X}$ in clean data.}. 

\begin{figure}[h]
    \centering
    \subfigure[]{\includegraphics[width=0.5\textwidth,height=0.25\textwidth]{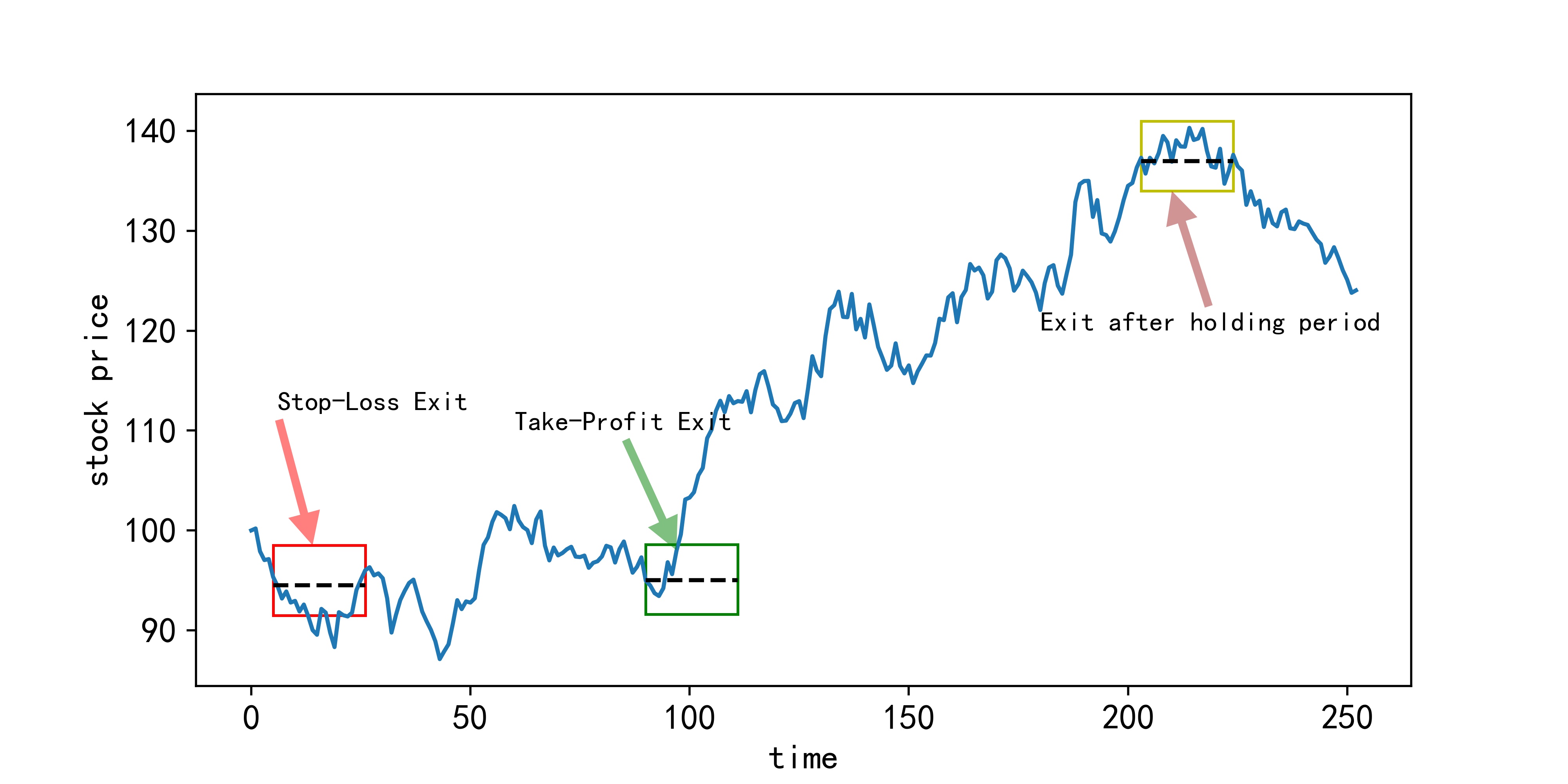}\label{fig: TRIPLE BARRIER METHOD}}
    \subfigure[]{\includegraphics[width=0.23\textwidth,height=0.23\textwidth]{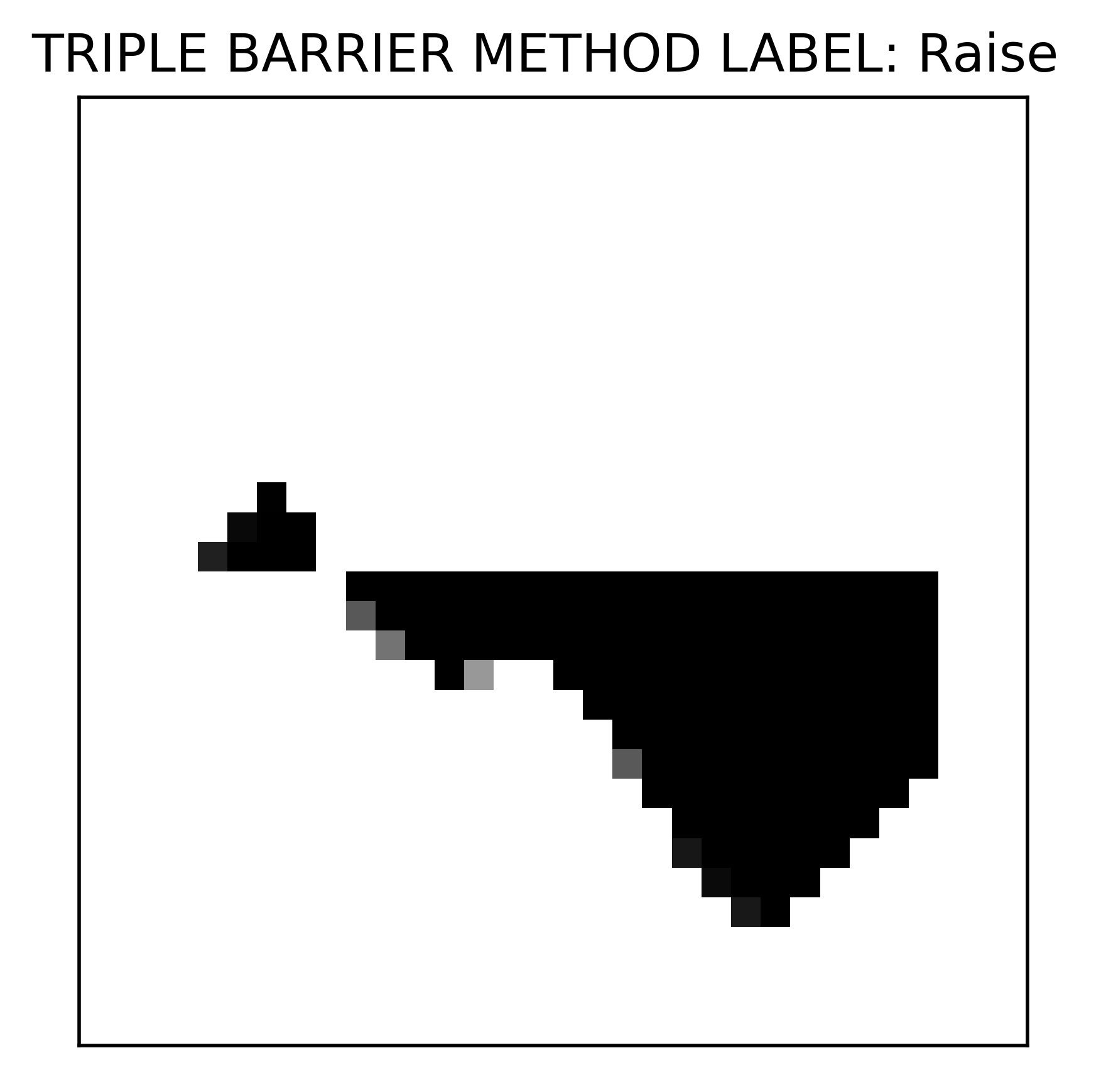}\label{fig: wrong label 2}}
    \subfigure[]{\includegraphics[width=0.23\textwidth,height=0.23\textwidth]{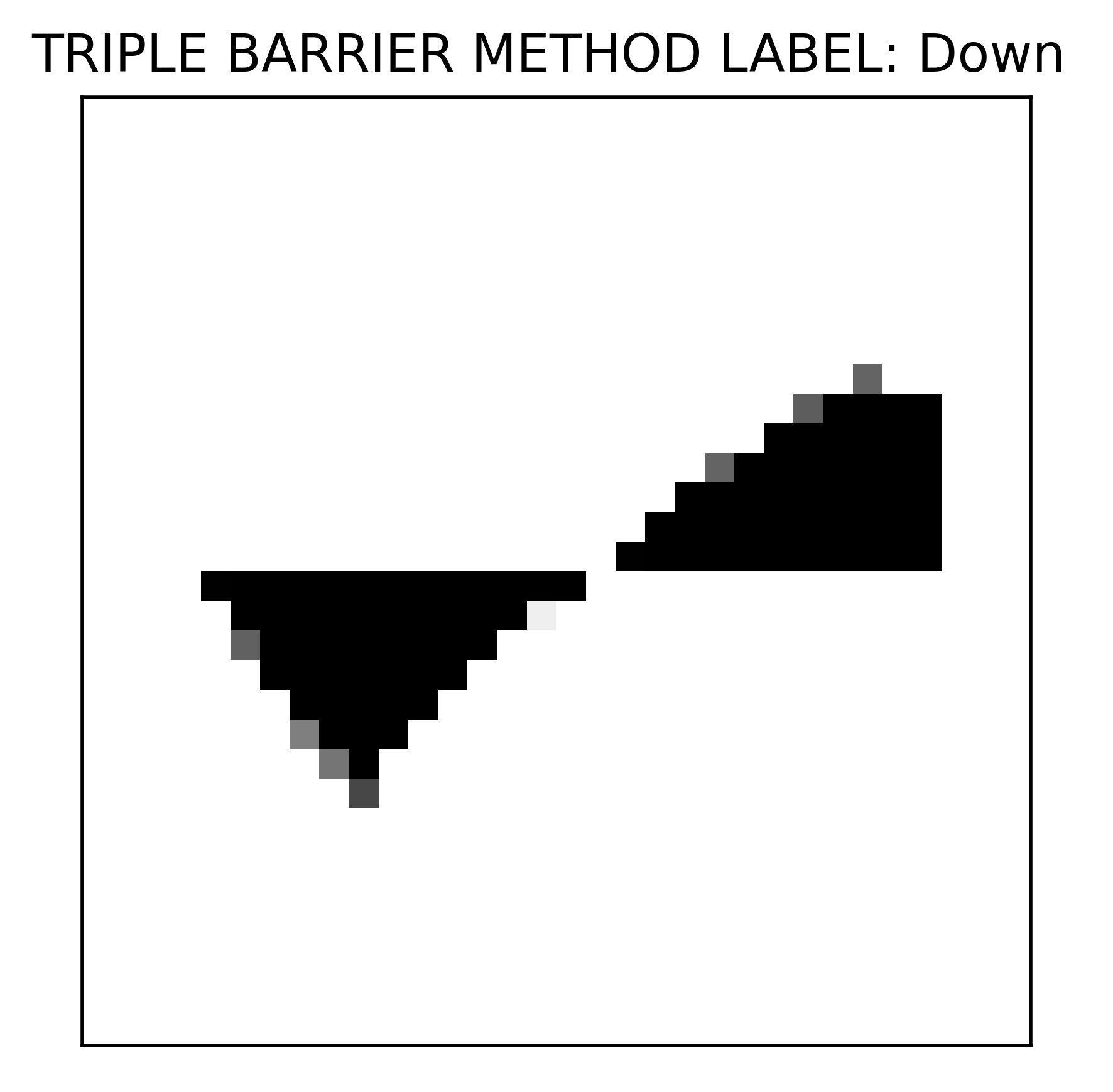}\label{fig: wrong label 1}}
    \caption{The triple barrier method with wrong label examples. (a). Manual labeling with triple barrier method, where the label of price trends is determined by the amount of change from beginning to the first stopping time that hits boundaries of the rectangular. (b). The stock goes down as a whole but the manual label is up. (c). The stock trend is uncertain in the whole picture but the manual label is down.}
    \label{fig:imageexample}
\end{figure}

For the second challenge, uncertain label quality mainly comes from the convectional labeling techniques, including the triple barrier approach, ensemble method and so on \citep{bounid2022advanced}.
The need of manual threshold setting, makes prediction results occasionally fall short of expectations (see Fig.\ref{fig:imageexample}). For example, we may get wrong labels based on the first passage time that hits the boundaries of the rectangular(Fig.\ref{fig:imageexample}(b)). Moreover, in some scenarios(Fig.\ref{fig:imageexample}(c)), it is difficult to artificially determine the trend with long prediction horizon. Hence, there are mainly two concerns for the human labeling ways: One is two artificial thresholds regarding the upper and lower bounds of the price; The other one is the window size chosen for calculating labels. For the former one, an adaptive threshold determined automatically by the financial market is more appropriate. And one way to solve this problem is through the meta learning framework, to learn the better label directly from its intrinsic pattern named ``Y" \footnote{ Y is also a unified representation that can be $\mathbf{Y}^{\prime}$ in noisy data.} For the latter issue, as different prediction horizons may give the label with completely different tendency directions, we need to design our model into a multi-task learning framework, where each task is distinguished by the length of its prediction horizon.

To summary,  we improve data information extraction shown in Fig.\ref{XY} with two aspects: historical time series data patterns (treated as features) and image patterns on the prediction horizon (treated as classification labels) \footnote{ We emphasize that sample in window size is referred to as historical data in this paper. Prediction horizon represents forecasting days. Here, the term ``samples" implies that original time series data has been divide into several sequences called samples through window size and prediction horizon.}. To achieve the goal of adaptive labeling, we make the following contributions:

$\bullet$ We propose better visualization methods to overcome  tendency confusion problems  for historical data and replace values of labels with patterns in the prediction horizon to obtain more Buy–Sell–Hold signals.

$\bullet$  We create a multi-task method to select the appropriate prediction horizon through dynamic forecast with a shared label correction model. 

$\bullet$ We develop a label correction model with two-branch neural network and without noisy real-world labels.
 
\section{ Related work}
\label{relatedwork}
In this section, we provide references to meta label correction, multi-task learning and bi-level optimization to guide us with tackling the challenges described in the introduction.

\subsection{ Meta label correction for label quality} 

To address label quality issue, some related research on label correction models has emerged as common approaches to handle the difficulties of high-quality label requirements. For instance, Zheng et al.\citep{zheng2021meta} propose a method called Meta Label Correction (MLC) under the framework of meta-learning. They construct two neural networks, one for classification prediction and the other for label correction via joint training with bi-level optimization. Their approach outperforms conventional re-weighting techniques in tasks concerning both language and images. Wu et al.\citep{wu2021learning} also utilize meta-learning, known as Meta Soft Label Corrector (MSLC), to modify labels automatically. Meanwhile, Mallem et al.\citep{mallem2023efficient} propose a new method dubbed CO-META with the help of multiple networks in the inner loop to reduce the risk of over-fitting and improve the quality of extracted features. Alongside the above label correction techniques with weak supervision, label generation is an alternative approach that predominantly relies on semi-supervised and self-supervised methodologies. Pham et al.\citep{pham2021meta} train a network called the teacher network to generate pseudo-labels on unlabeled data. Teacher network can instruct another network called the student network under the framework of semi-supervised learning. Ma et al.\citep{ma2022denoised} take label generation as the pretext task with a self-supervised method and focus on obtaining better financial labels.

\subsection{Multi-task learning for financial data}

Moreover, there are also several references regarding multi-task learning in the field of financial data. Park et al.\citep{park2022stock} propose an ensemble model established by a long short-term memory model integrated with random forest to predict stock market returns and classify return directions based on multi-task learning. Chandra and Cripps\citep{chandra2018bayesian} regard different window sizes in time series as tasks and utilize Bayesian inference and multi-task learning to obtain predictions of financial data. Driven by these, we intend to get the optimal performance of all the possibilities of various forecasting days through a shared label correction model via multi-task learning and meta learning.

\subsection{Bi-level optimization for gradient updating}

Each of the aforementioned label correction methods involves bi-level optimization, resulting in several difficulties in gradient updating and computational memory management. Explicit and implicit updating rules are two different approaches for gradient-based optimizations. For the former case, Gao et al.\citep{gao2022value} utilize the value function approach to solve tractable convex sub-problems and develop a theoretical framework for sequential convergence towards stationary solutions. As the assumption of lower-level convexity is too restrictive, Liu et al.\citep{liu2021towards} propose an algorithm called Initialization Auxiliary and Pessimistic Trajectory Truncated Gradient Method (IAPTT-GM) to handle large-scale and non-convex bi-level optimization problems with a first-order explicit gradient method and dynamic initialization. Beyond first-order gradient methods, second-order gradient techniques that entail hyper-gradient computation at the upper level are also prevalent in optimization literature. To avoid the expensive computational cost for the Hessian matrix, Liu et al.\citep{liu2018darts} use the finite difference approximation to reduce its complexity. As for the latter case, implicit gradient update methods are also increasingly garnering attentions. Rajeswaran et al.\citep{rajeswaran2019meta} establish an implicit MAML (iMAML) algorithm based on the implicit function theorem to find an approximate solution to the inner-level optimization problem without storing or differentiating the inner optimization path. Zhang et al.\citep{zhang2021idarts} also apply the implicit function theorem to address the hyper-gradient computation of Differentiable Architecture Search (DARTS).

Although prior research has demonstrated connections between meta label correction, multi-task learning and bi-level optimization, it has not been implemented in the financial domain. We desire to automatically choose the right window size as well as the adaptive upper/lower bound for the label of future trend, which is more feasible in real world applications. Hence, we develop an automatic label correction model with the help of a small portion of clean labels.

\section{\textbf{Meta Label Correction}}
\label{metalabelcorrection}
High-quality labels, unlike noisy ones, enable the significant improvement of neural network performance. Currently, many studies focus on  public benchmark datasets such as  CIFAR-10,  CIFAR-100 \citep{krizhevsky2009learning}, MNIST \citep{lecun1998gradient}, Clothing1M \citep{xiao2015learning}, etc. However, accurately labeled financial data also plays an essential role in predicting stock trends. In this section, we introduce data visualization methods and our proposed model.

\subsection{\textbf{Data Processing}}\label{AA}
We divide the training dataset into two parts: a small amount of data with clean labels called \emph{meta dataset}, and a large amount of data with noisy labels called \emph{noisy dataset}. Moreover, in order to show Buy–Sell–Hold signals more intuitively, the visualization method of prediction horizon is different from the approach of historical data.

\begin{figure}[htbp]
\centerline{\includegraphics[width=0.7\linewidth]{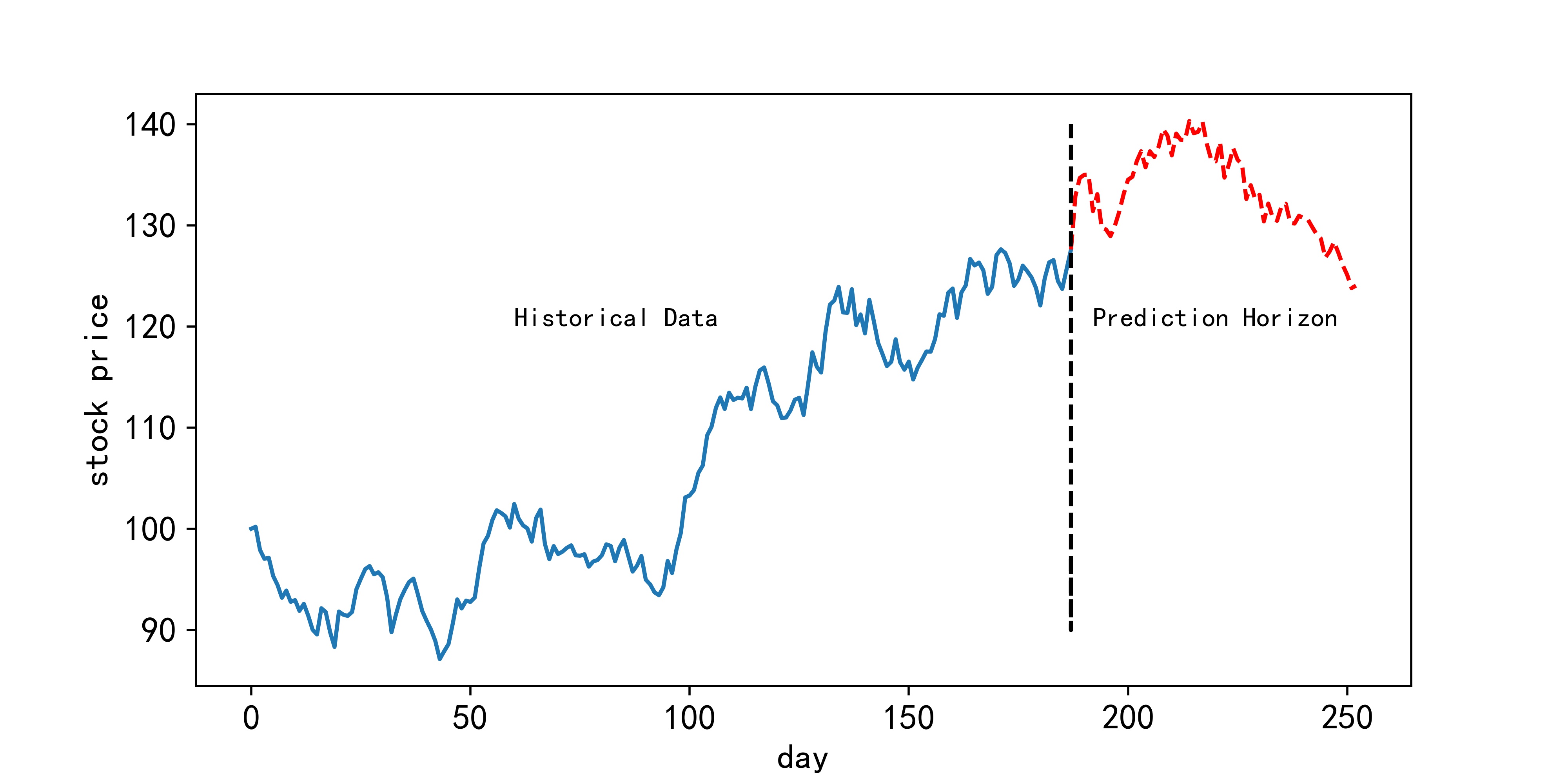}}
\caption{ Historical data is used for generating samples of training features denoted as ``X" in the following algorithm. Prediction horizon is used for generating labels, whose image patterns are denoted by ``Y" in the following.}
\label{XY}
\end{figure}
 
\subsubsection{\textbf{Historical data visualization method}} 
\label{historicalimage}
\begin{figure}[htbp]
\centerline{\includegraphics[width=0.65\linewidth]{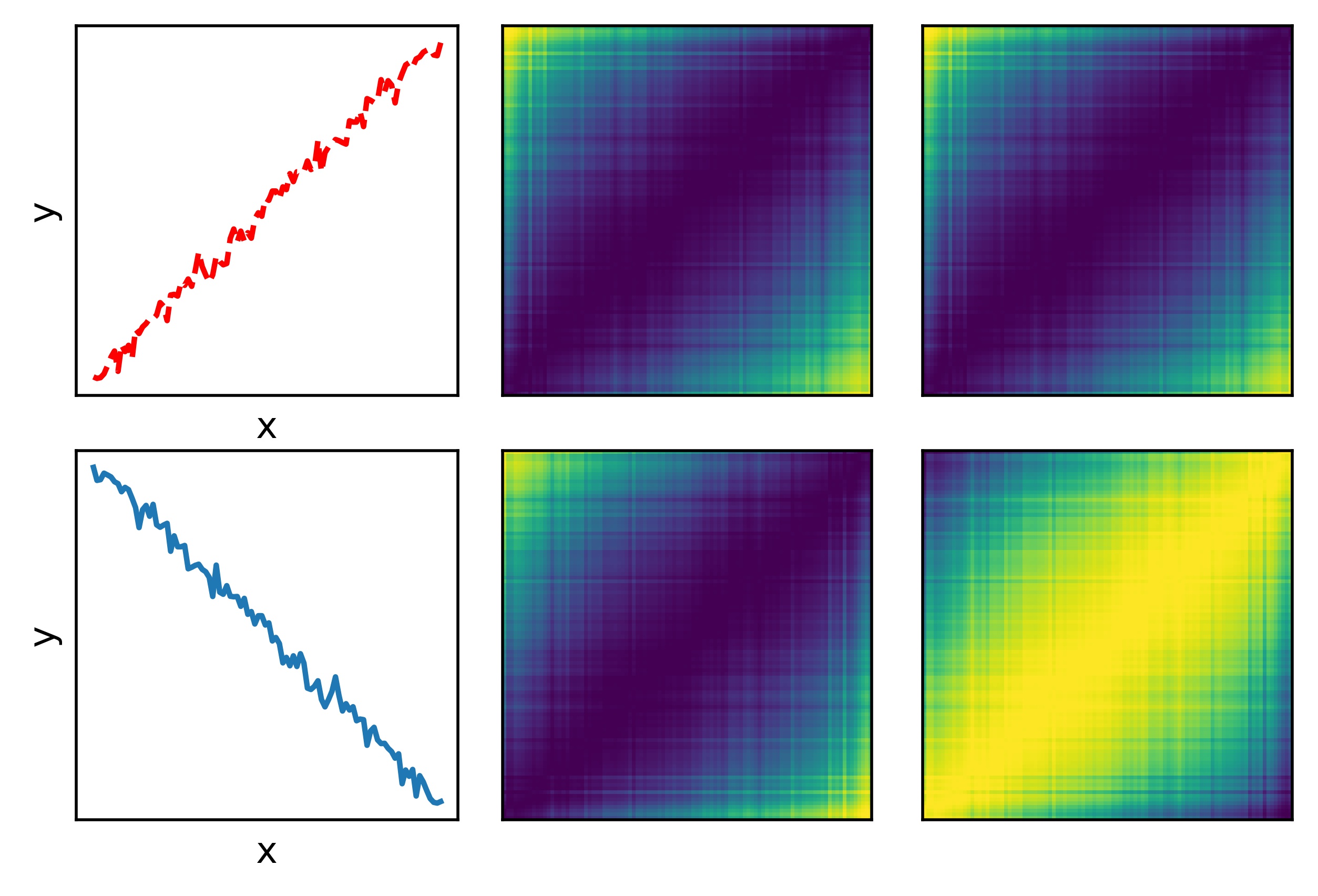}}
\caption{Tendency confusion problem. The upper left is $y=2x+N(0,0.5)$. $N(0,0.5)$ is a Gaussian noise. The upper middle is the corresponding GAF of $y=2x+N(0,0.5)$. The lower left is $y=2x+N(0,0.5)$ in reverse order. The lower middle is the corresponding GAF plot of reverse data. The right column with our proposed method SGAF, shows different image patterns of the ascending and descending trends respectively, while the middle column (original GAF) has tendency confusion with the same patterns for different trends.}
\label{gaf}
\end{figure}

The traditional methods we used for stock prediction, such as the LDE-Net model \citep{yang2023neural}, often focus on univariate time series with the limit of capturing potential correlations among data. As the effectiveness of GAF in financial prediction \citep{chen2016financial}, we adopt it and obtain the Gramian Matrix in Eq.~\eqref{GM}
\begin{equation}\label{GM}
\begin{aligned}
& \qquad G_{k}=\left[\begin{array}{ccc}
\langle \tilde{x}_{1+k}, \tilde{x}_{1+k} \rangle  & \cdots & \langle \tilde{x}_{1+k}, \tilde{x}_{n+k}\rangle \\
\vdots & \ddots & \vdots \\
\langle \tilde{x}_{n+k}, \tilde{x}_{1+k}\rangle & \cdots & \langle \tilde{x}_{n+k}, \tilde{x}_{n+k}\rangle
\end{array}\right], 
\end{aligned}
\end{equation}

where  $\langle\tilde{x}_{i+k},\tilde{x}_{j+k}\rangle=\tilde{x}_{i+k}\cdot \tilde{x}_{j+k}-\sqrt{1-\tilde{x}_{i+k}^2} \cdot \sqrt{1-\tilde{x}_{j+k}^2}$ and $\tilde{x}_{i+k}$ or $\tilde{x}_{j+k}$ in $G_{k}$ represents a certain scaling value in $\mathrm{X_{k}}=\left\{x_{1+k}, x_{2+k}, \ldots, x_{n+k}\right\}$, which is the $k$th sample of train data $\mathrm{X}$ divided by window size $n$. Furthermore, $i=1,\ldots,n$, $j=1,\ldots,n$ and $k=0,\ldots,N-n-(H-1)$. Moreover, train data $X$ with the number of time series data $N$ and the length of prediction horizon $H$ can be written as $\mathrm{X}=\left\{x_1, \ldots, x_n, x_{n+1}, \ldots, x_{n+H}, \ldots, x_{N}\right\} $. The design of this transformation method retains original information through upper and left parts of the matrix and captures additional hidden information through the inner product $\langle \cdot \rangle$. However, as demonstrated in Eq.\eqref{GM} and Fig.\ref{gaf}, the GAF method suffers from a tendency to confuse different patterns.

To address this issue, we propose a novel method named Sign function multiplied with Gramian Matrix (SGAF) through adding a plus-minus sign in front of Gramian Matrix and demonstrate effectiveness in our experiments. Mathematically, the sign function is given by 
\begin{equation}\label{sign}
\begin{aligned}
sign(G_{k})=\begin{cases}1, 
  & \text{ if } \frac{x_{n+k}-x_{1+k}}{n-1}\ge 0,\\
  -1, & \text{ if } \frac{x_{n+k}-x_{1+k}}{n-1}<  0.
\end{cases}
\end{aligned}
\end{equation}
Then, after splitting time series, the modified GAF matrix of $k$th sample used for generating an image becomes
\begin{equation}\label{mgaf}
\begin{aligned}
\widetilde{G_{k}} = sign(G_{k})\cdot G_{k},
\end{aligned}
\end{equation}

where $k=0,\ldots,N-n-(H-1)$, with $N, n, H$ following their previously established definitions. While retaining original characteristics of the GAF matrix, this construction can aid in identifying the sequence's overall trend (see Fig.~\ref{gaf}).

On the other hand, RP is another visualization tool for studying financial data \citep{addo2013nonlinear}. Different from correlation-based method GAF, an image matrix of RP is generated by the distance among reconstructed points \citep{addo2013nonlinear}. However, this method also faces the challenge of tendency confusion. Therefore, we also modify the RP method by adding a plus-minus sign function in front of RP Matrix, named Sign function multiplied with RP Matrix (SRP). Following \citep{addo2013nonlinear}, the embedding dimension $d =1$ and time delay $\tau =1$ for financial data. Hence, each element of the binary recurrence matrix $R_{k}$ for the $k$th sample of train data $\mathrm{X}$ can be defined as
\begin{equation}
R_{k}(i, j)= \begin{cases}1 & \text { if }\|x_{i+k}-x_{j+k}\| \leq \varepsilon \\ 0 & \text { otherwise, }\end{cases}
\label{rpmatrix}
\end{equation}
where $\|\cdot\|$ is a norm and $\varepsilon$ is the recurrence threshold. The window size of $k$th sample equals $n$, similar to the above GAF method, so that we have $i=1,\ldots,n$, $j=1,\ldots,n$ and $k=0,\ldots,N-n-(H-1)$. Here, $N$ and $H$ are also the total number of data and the length of prediction horizon. If $R_{k}(i, j)=1$, RP visualizes the matrix $R_{k}$ with a black point at coordinates $(i, j)$. Based on the rule of thumb \citep{he2020global}, we set the percentage of black points as $10\%$. The threshold $\varepsilon$ is computed such that $10\%$ of the points are smaller than $\varepsilon$. Then, a modified RP matrix for the $k$th sample is defined as
\begin{equation}
\label{mrp}
\begin{aligned}
\widetilde{R_{k}} = sign(G_{k})\cdot R_{k}.
\end{aligned}
\end{equation}

Here, $sign(G)$ is the same as it in Eq.~\eqref{sign}. Consequently, we are able to extract additional features from time series itself with the assistance of these two proposed visualization methods, SGAF and SRP. Accordingly, patterns generated by SGAF and SRP are all called ``X".

\begin{figure}[htbp]
\centerline{\includegraphics[width=0.65\linewidth]{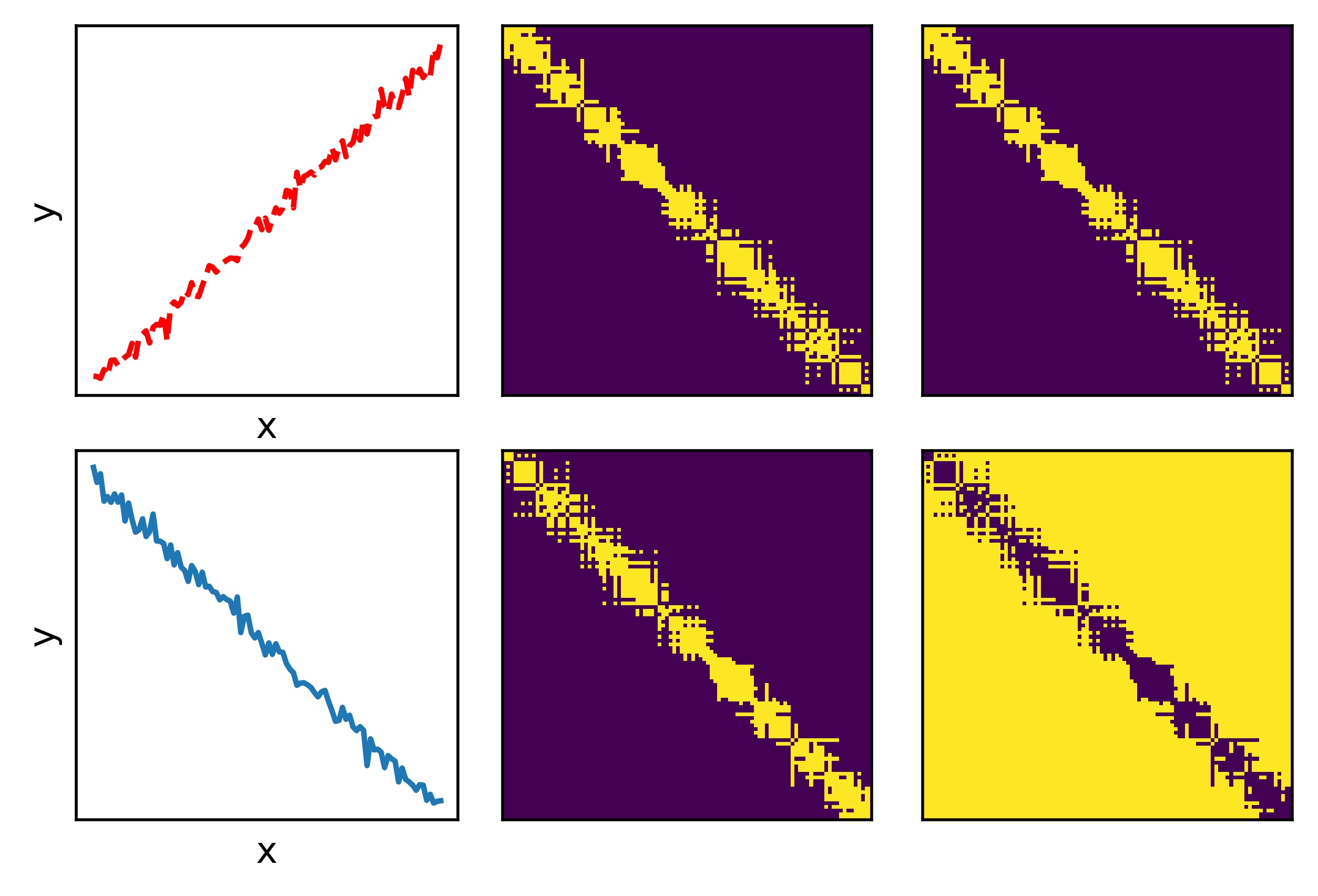}}
\caption{ An example of the SRP method. (The upper left is $y=2x+N(0,0.5)$ and lower left is $y=-2x-N(0,0.5)$, where $N(0,0.5)$ is the Gaussian noise. The middle plots are using corresponding RP method on the left column. The right column are images with our SRP method, with a sign function added into RP.}

\label{gafrp}
\end{figure}

\subsubsection{\textbf{Prediction horizon visualization method}}
\label{horizonimage}
\begin{figure}[htbp]
\centerline{\includegraphics[width=0.7\linewidth]{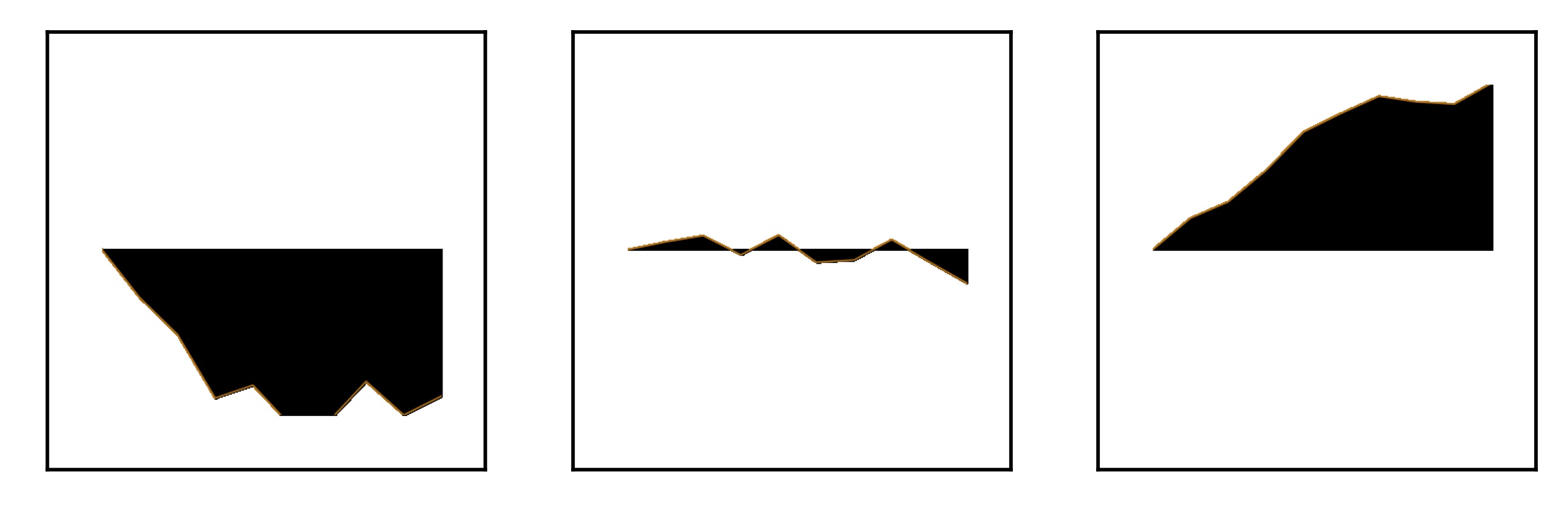}}
\caption{ A particular case with RRP. The left figure shows that $x_{n+k+h}-x_{n+k} <0$ with a ``fall" label. The middle figure shows that the fluctuation of $\|x_{n+k+h}-x_{n+k}\|$ is small that implies the sequence is stationary. The right figure is the opposite of the left figure with  $x_{n+k+h}-x_{n+k} >0$ indicating a rise. }
\label{rrp}
\end{figure}

Different from traditional classification methods, we substitute values of labels with patterns called ``Y" in the forecasting horizon, which assists in obtaining more information on labels to capture Buy–Sell–Hold signals. To categorize the tendency into three categories: rise, stationary, and fall, we propose a novel transformation technique called Relative Ratio Plot (RRP, see  Fig.~\ref{rrp}). The relative ratio is calculated through Eq.~\eqref{RR},
\begin{equation}\label{RR}
\begin{aligned}
    ratio_{k}(h) = \frac{x_{n+k+h}}{x_{n+k}}-1, \quad \text{for}\; k=0,\dots,N-n-(H-1); h=1,\dots,H,
\end{aligned}
\end{equation}
where $H$ is the length of the prediction horizon and $n$ is the window size of training data. Here, $k$ represents the $k$th sample in horizon prediction after splitting data. The specific values of $n, N, H$ can be shown in Section \ref{datasets}. For each the $k$th sample, we utilize $\{ratio_{k}(1), ratio_{k}(2),\dots, ratio_{k}(H)\}$ to generate an image with replace of a label's value (see  Fig.~\ref{rrp}). In a nutshell, we provide a pseudo code in \ref{A} (see Algorithm \ref{geimage}) to present the process of generating images ``X" and ``Y" respectively. 

\subsubsection{\textbf{Labeling method}} 
For the prediction horizon, we employ Triple Barrier Method \citep{bounid2022advanced}, a conventional technique, to label it. However, since this approach requires the manual definition of thresholds, it is not always reliable, especially in noisy data (see Fig.\ref{fig:imageexample}). To obtain clean labels, we provide a benchmark (see Eq.~\eqref{rb}) based on \citep{li2022selective, lai2009evolving}. 

\begin{equation}\label{rb}
baseline = \omega * \left(\frac{(1+rate)^{H+1}-(1+rate)}{rate\times H} - 1 \right).
\end{equation}

Here, $\omega$ is a constant adjusted by stock price value, $rate = 0.005$ and $H$ is the forecasting time period, according to \citep{li2022selective,liu2016encoding}. The approach of labeling the prediction horizon has the following mathematical expression:
\begin{equation}\label{label1}
\begin{cases}
 &\frac{1}{H} {\textstyle \sum_{h=1}^{H} ratio_{k}(h)} < -baseline \quad\quad \quad\quad\quad\quad\quad \text{ label: fall}  \\
  & -baseline < \frac{1}{H} {\textstyle \sum_{h=1}^{H} ratio_{k}(h)} <baseline \quad\quad \text{ label: stationary}\\
  & \frac{1}{H} {\textstyle \sum_{h=1}^{H} ratio_{k}(h)} > baseline    \quad\quad\quad\quad\quad\quad\quad\quad \text{label: rise}
\end{cases}
\end{equation}

\begin{figure}[htbp]
\centerline{\includegraphics[width=0.5\linewidth]{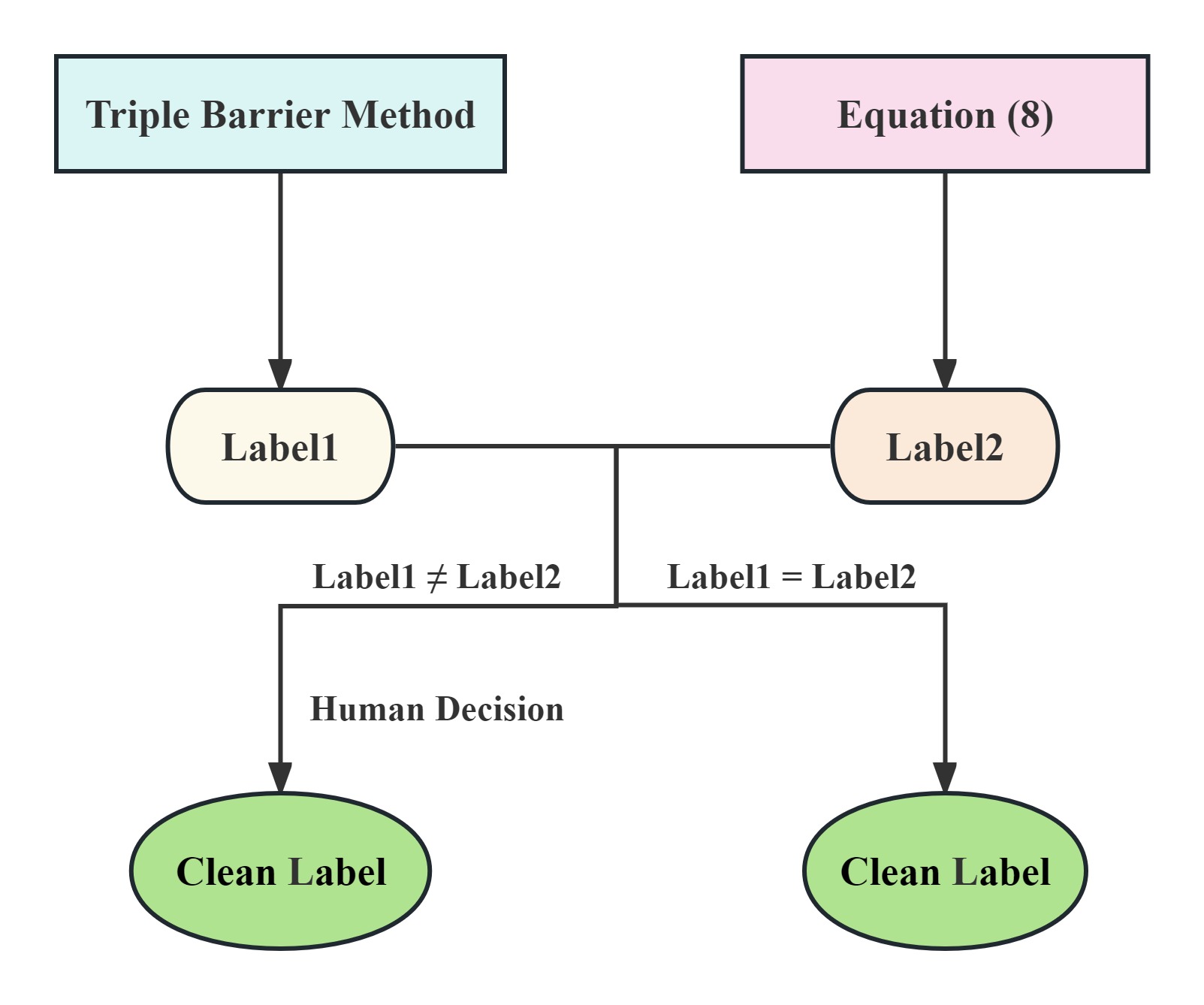}}
\caption{Flow chart for obtaining clean labels. If labels gained by this two methods are the same, we finally regard it as a clean label. If they are different, we manually label them due to the small number of data points.}
\label{cleanlabel}
\end{figure}

In order to ensure the accuracy of clean labels, we compare labels obtained by Eq.~\eqref{label1} with labels obtained by the Triple Barrier Method to get final clean labels (See Fig.~\ref{cleanlabel}).

\subsection{\textbf{Meta learning for label correction}}\label{BB}
Given the high cost of collecting large numbers of clean labels, we propose a multi-task meta-learning method label correction called \textbf{Multi-task Meta Label Correction} (MMLC) for financial data. We create a framework wherein a label correction network (LCN) is considered as \emph{meta model}  and classification networks are regarded as \emph{main models}.

\subsubsection{\textbf{Label correction formulation based on meta learning and multi-task learning}}\label{labelcorrection}
In our framework, we provide meta-training tasks $\left\{\mathcal{T}_i\right\}_{i=1}^I$ obtained from $P(\mathcal{T})$ with $I$ as the number of tasks. Here, $P(\mathcal{T})$ implies the distribution over tasks and each task $\mathcal{T}_i$ is split in accordance with the length of prediction horizon $H$. Hence, based on training data for each stock, we can obtain clean data samples $\mathcal{D}_i=\{\mathbf{X},\mathbf{y}\}^{m}$ and noisy data samples $\mathcal{D'}_i =\{\mathbf{X'},\mathbf{Y'}\}^{M}$, where $m \ll M$. Here, $\mathbf{X} \in \mathcal{X}$ and $\mathbf{y} \in \mathcal{Y}$ denote the image data of historical data in the clean domain and their corresponding clean labels, while $\mathbf{X'} \in \mathcal{X'}$ and $\mathbf{Y'} \in \mathcal{Y'}$ denote the image data of historical data and prediction horizon in the noisy domain (See Fig.\ref{XY}). The image generations of $\mathbf{X'},\mathbf{Y'}$ and $\mathbf{X} $ are shown in Sections \ref{historicalimage} and \ref{horizonimage}, which can also be found in Fig.~\ref{dataset}.

\begin{figure}[htbp]
\centerline{\includegraphics[width=0.4\textwidth,height=0.5\textwidth]{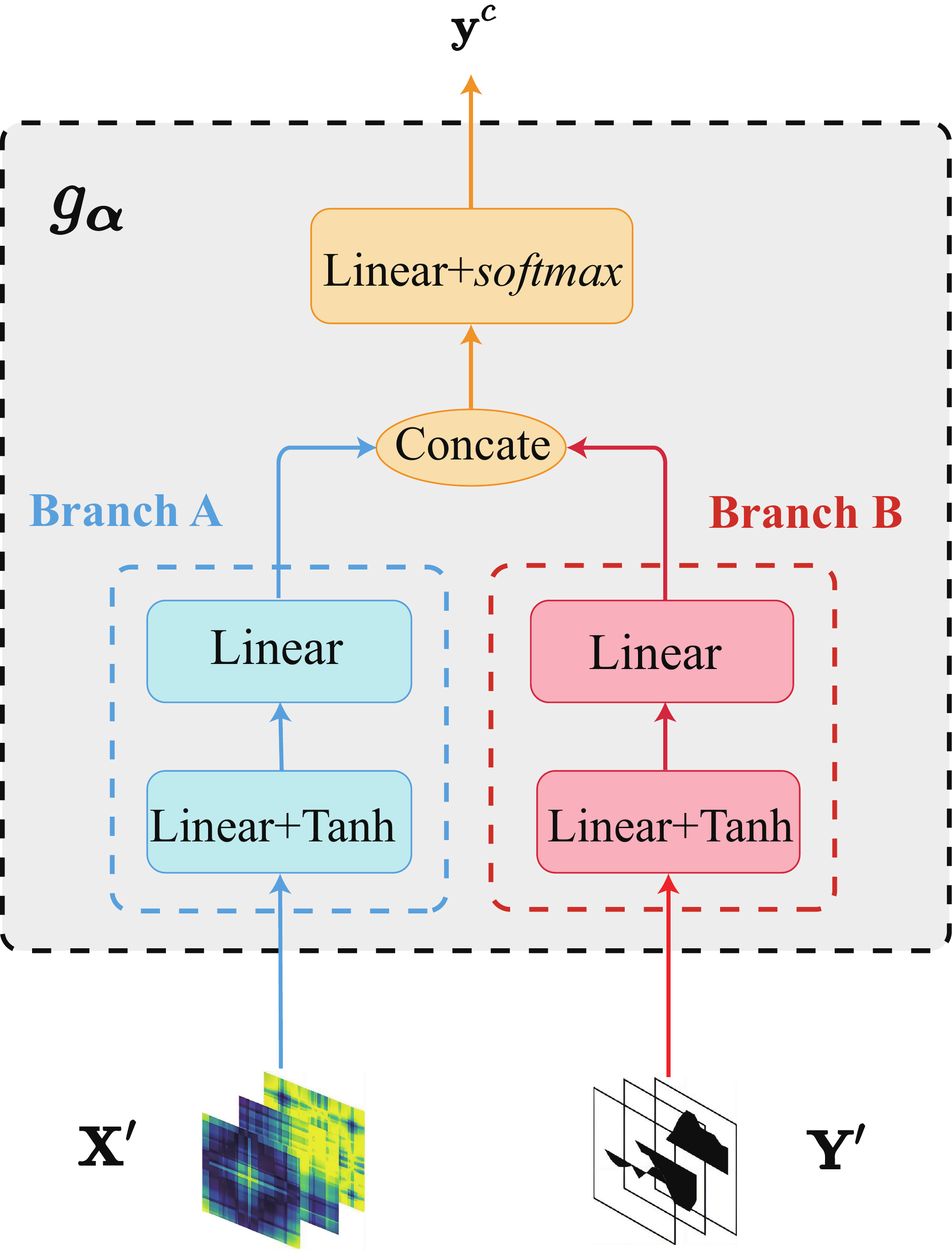}}
\caption{The label correction network (LCN) architecture $g_{\boldsymbol{\alpha}}\left(\mathbf{X}^{\prime}, \mathbf{Y}^{\prime}\right)$, the inputs of which are noisy examples $\{\mathbf{X'},\mathbf{Y'}\}$. The output of this \emph{meta model} is the corrected label $\mathbf{y}^{c}$.}
\label{metamodel}
\end{figure}
 
First, we present our label correction method in which working on a fixed task $\mathcal{T}_i$. We attempt to utilize \emph{meta model} to generate corrected labels $\mathbf{y}^{c}$ with $\{\mathbf{X}^{\prime},\mathbf{Y}^{\prime}\}$ without the interference of noisy labels. The structure of \emph{meta model} (see Fig.~\ref{metamodel}) is composed of two branch neural networks and a generation module that resemble Pseudo‐Siamese network \citep{jiang2021pseudo} with a smaller size. Simultaneously, the parameters of \emph{meta model} $\boldsymbol{\alpha}$ are regraded as \emph{meta-knowledge} so that \emph{meta model} LCN can be denoted by $g_{\boldsymbol{\alpha}}\left(\mathbf{X}^{\prime}, \mathbf{Y}^{\prime}\right)$. Here, one thing needs attention. The parameters $\boldsymbol{\alpha}$ of \emph{meta model} are optimized with the help of labels of clean data $\mathcal{D}_i=\{\mathbf{X},\mathbf{y}\}^{m}$ and without the influence of noisy data. Moreover, for each task $\mathcal{T}_i$, \emph{main model} $f_{\mathbf{w}_{i}}$ is used for prediction after training with task-specific parameters $\mathbf{w}_{i}$ and is independent of the other tasks. Therefore, the goal of task $\mathcal{T}_i$ is to find a good \emph{meta-knowledge} $\boldsymbol{\alpha}$ to generalize across all tasks and to learn task-specific parameters $\mathbf{w}_{i}$ to minimize the population loss.

\begin{figure}[htbp]
\centerline{\includegraphics[width=0.8\linewidth]{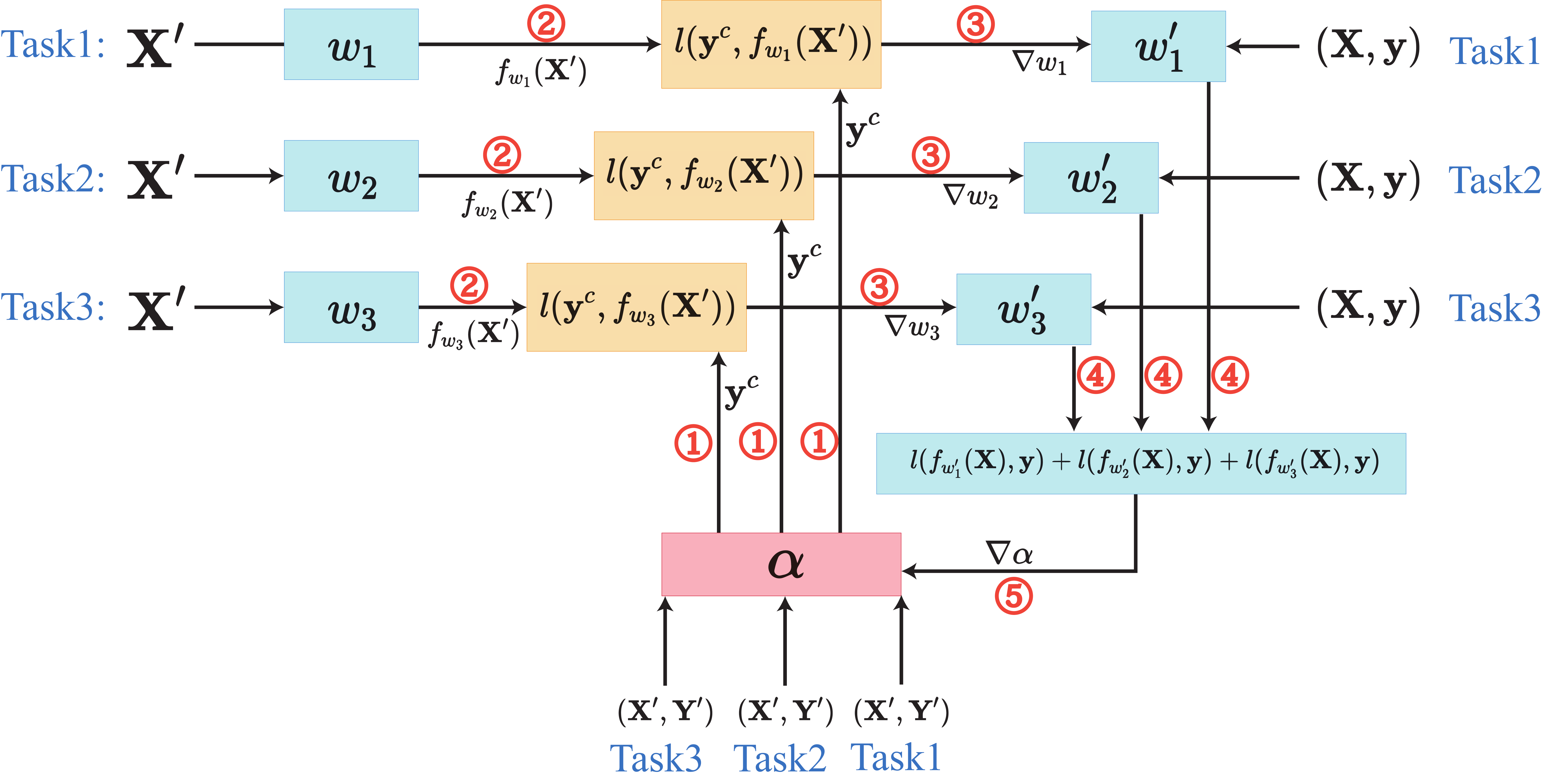}}
\caption{Block diagram of the proposed MMLC method. Here, we only show three tasks. Each task has an independent \emph{main model} $f_{w_i}$ with parameters $w_i, (i=1,2,3)$ and a common \emph{meta model} (see Fig.~\ref{metamodel}) with parameters $\alpha$. In each task $\mathcal{T}_i \sim P(\mathcal{T})$, $(\mathbf{X}^{\prime},\mathbf{Y}^{\prime})$ and $(\mathbf{X},\mathbf{y})$ are from noisy and clean samples, respectively. }
\label{MLC}
\end{figure}

The training process, depicted in Fig.~\ref{MLC}, unfolds as follows: \textcolor{red}{\textcircled{1}} From tasks $\mathcal{T}_i \sim P(\mathcal{T}), i = 1\cdots I$, we select noisy samples $(\mathbf{X}^{\prime},\mathbf{Y}^{\prime})$ and clean examples $(\mathbf{X},\mathbf{y})$. The noisy examples are processed by the LCN to obtain corrected labels. \textcolor{red}{\textcircled{2}} The noisy data $\mathbf{X}^{\prime}$ is then used as input for the current task’s classifier to generate predictions. \textcolor{red}{\textcircled{3}} The classifier’s parameters are updated based on the loss calculated between the corrected and predicted labels. \textcolor{red}{\textcircled{4}} Clean example pairs $(\mathbf{X},\mathbf{y})$ are fed into the classifier to compute the classification loss. \textcolor{red}{\textcircled{5}} This process is repeated across all tasks, accumulating individual classification losses. The total classification loss is then used to update the LCN parameters through gradient descent. Furthermore, bi-level optimization links the two networks, as formulated below:

\begin{equation}
\begin{aligned}
& \underset{\boldsymbol{\alpha}}{\min } \sum_{i=1}^I \mathbb{E}_{(\mathbf{X}, \mathbf{y}) \in D_{i}} \ell\left(\mathbf{y}, f_{\mathbf{w}_{i}^*(\boldsymbol{\alpha})}(\mathbf{X})\right) \\
 \text { s.t. } \quad & \mathbf{w}_{i}^*(\boldsymbol{\alpha})=\underset{\mathbf{w}_{i}}{\arg \min } \mathbb{E}_{\left(\mathbf{X}^{\prime}, \mathbf{Y}^{\prime}\right) \in D_{i}^{\prime}}\ell\left(g_{\boldsymbol{\alpha}}\left(\mathbf{X}^{\prime}, \mathbf{Y}^{\prime}\right), f_{\mathbf{w}_{i}}(\mathbf{X}^{\prime})\right),
\end{aligned}
\label{bilevel}
\end{equation}
where $I$ is the number of tasks and $\ell(\cdot)$ is cross entropy loss for classification. In this bi-level optimization problem, whenever \emph{meta-knowledge} $\boldsymbol{\alpha}$ updates, the optimal $\mathbf{w}_{i}^*$ with each task $\mathcal{T}_i$ is required.

\subsubsection{\textbf{Gradient-based bi-level optimization}}\label{gtadient}

In bi-level optimization, it is necessary to update the parameters in \emph{main model} and \emph{meta model} with two loops. For the lower-level (inner-loop)\footnote{We refer to the upper-level and lower-level in the bi-level framework as the outer-loop and inner-loop in the meta-learning.}, we select K-step SGD \citep{finn2017model} approximation to obtain the optimal main model for each task $\mathcal{T}_i$ given a $\boldsymbol{\alpha}$. Furthermore, we only use the last step of the SGD update in each task of the \emph{main model} (see Eq.~\eqref{mainopt}) for updating the \emph{meta-knowledge} $\boldsymbol{\alpha}$. 
\begin{equation}
\mathbf{w}_{i}^*(\boldsymbol{\alpha}) \approx \mathbf{w}^{\prime}_{i}(\boldsymbol{\alpha})=\mathbf{w}_{i}-\eta \nabla_{\mathbf{w}_{i}} \mathcal{L}_{D^{\prime}_{i}}(\boldsymbol{\alpha}, \mathbf{w}_{i}).
\label{mainopt}
\end{equation}

Here, $\mathbf{w}^{\prime}_{i}(\boldsymbol{\alpha})\triangleq\mathbf{w}_{i,K}(\boldsymbol{\alpha})$ is the parameters after the last step update in task $\mathcal{T}_i$.  Also, $\nabla_{\mathbf{w}_{i}} \mathcal{L}_{D^{\prime}_{i}}(\boldsymbol{\alpha},\mathbf{w}_{i})\triangleq\nabla_{\mathbf{w}_{i,K-1}}\mathbb{E}_{\left(\mathbf{X}^{\prime}, \mathbf{Y}^{\prime}\right) \in D_{i}^{\prime}}\ell\left(g_{\boldsymbol{\alpha}}\left(\mathbf{X}^{\prime}, \mathbf{Y}^{\prime}\right), f_{\mathbf{w}_{i,K-1}}(\mathbf{X}^{\prime})\right)$ is the lower-level loss function and $\mathbf{w}_{i}\triangleq\mathbf{w}_{i,K-1}$ is the parameters after the penultimate update. The learning rate $\eta$ in each \emph{main model} is assumed to be the same. 

\begin{algorithm}
\caption{\emph{Main Model} Gradient Computation}\label{inneral}
\footnotesize{
\KwIn{Task $\mathcal{T}_i$, \emph{meta-knowledge} $\boldsymbol{\alpha}$, learning rate $\eta$, noisy dataset $\mathcal{D'}_i =\{\mathbf{X'},\mathbf{Y'}\}^{M}$,\emph{meta model} $g_{\boldsymbol{\alpha}}$ }
\KwOut{\emph{Main model} parameters $\mathbf{w}^{\prime}_{i}(\boldsymbol{\alpha}) = \mathbf{w}_{i,K}$ }
Initialize \emph{main model} parameters $\mathbf{w}_{i,0}$.\\
\For{$k=0$ to $K-1$}{
$\{\mathbf{X'},\mathbf{Y'}\} \leftarrow \text{SampleMiniBatch}(\mathcal{D'}_i,n)$.\\
\text{Update main model parameters} $\mathbf{w}_{i,k+1} = \mathbf{w}_{i,k} - \eta \nabla_{\mathbf{w}_{i,k}} \mathcal{L}_{D^{\prime}_{i}}(\boldsymbol{\alpha}, \mathbf{w}_{i,k}) $\\
}
}
\end{algorithm}

For the upper-level (outer-loop), it is inevitable to calculate $\frac{\partial \mathbf{w}_{i}^*(\boldsymbol{\alpha})}{\partial \boldsymbol{\alpha}}$ that involves second-order gradient computation \citep{chen2022gradient}. We select a method called meta-parameter with K-step SGD from main-parameters in \citep{zheng2021meta} to update our \emph{meta model}. But, different from it, we calculate the Hessian matrix $\frac{\partial^2}{\partial \mathbf{w_i} \partial \mathbf{w_i}} \mathcal{L}_{D_{i}^{\prime}}(\boldsymbol{\alpha}, \mathbf{w_i})$ with the Taylor expansion method mentioned in \citep{liu2018darts}. Compared with Identity matrix used in \citep{zheng2021meta}, our method of computing the approximated Hessian matrix can achieve better performance, in terms of accuracy and F-1 score (see Section \ref{results}). In contrast to explicitly calculating the Hessian matrix, we decrease the complexity from $O( \|\mathbf{w}\| \cdot \|\mathbf{w}\| )$ to $O(\|\mathbf{w}\|)$ \citep{liu2018darts}. Denote the meta loss as $\sum_{i=1}^I\mathcal{L}_{D_{i}}\left(\mathbf{w}_{i}^{*}(\boldsymbol{\alpha})\right)\triangleq\sum_{i=1}^I \mathbb{E}_{(\mathbf{X}, y) \in D_{i}} \ell\left(y, f_{\mathbf{w}_{i}^{*}(\boldsymbol{\alpha})}(\mathbf{X})\right)$. Then, we have
\begin{equation}
\begin{aligned}
\min _{\boldsymbol{\alpha}} \sum_{i=1}^I\mathcal{L}_{D_{i}}\left(\mathbf{w}_{i}^{*}(\boldsymbol{\alpha})\right) &\approx \sum_{i=1}^I\mathcal{L}_{D_{i}}\left(\mathbf{w}^{\prime}_{i}(\boldsymbol{\alpha})\right)\\
&=\sum_{i=1}^I\mathcal{L}_{D_{i}}\left(\mathbf{w}_{i}-\eta \nabla_{\mathbf{w}_{i}} \mathcal{L}_{D^{\prime}_{i}}(\boldsymbol{\alpha}, \mathbf{w}_{i})\right).
\end{aligned}
\label{metaopt}
\end{equation}
Subsequently, the update rule of \emph{meta-knowledge} $\boldsymbol{\alpha}$ can be written with the learning rate $\mu$ as follows:
\begin{equation}
\boldsymbol{\alpha}=\boldsymbol{\alpha}-\mu \nabla_{\boldsymbol{\alpha}}\sum_{i=1}^I \mathcal{L}_{D_{i}}\left(\mathbf{w}^{\prime}_{i}(\boldsymbol{\alpha})\right).
\label{metaknowledge}
\end{equation}
Then, the meta-parameter gradient from previous $T$ steps in upper-level is shown as\footnote{Because of linearity, $\sum_{i=1}^N$ and $\nabla_{\boldsymbol{\alpha}}$ can be switched.}:
\begin{equation}
\begin{split}
\frac{\partial \mathcal{L}_{D_{i}}\left(\mathbf{w}^{\prime}_{i}(\boldsymbol{\alpha})\right)}{\partial \alpha} 
=& g_{\mathbf{w}_{i}^{\prime}}(I-\eta \nabla_{\mathbf{w}_{i},\mathbf{w}_{i}}\mathcal{L}_{D^{\prime}_{i}}(\boldsymbol{\alpha}, \mathbf{w}_{i})) \frac{g_{\mathbf{w}_{i}^{\top}}}{\left\|g_{\mathbf{w}_{i}}\right\|^2}\frac{\partial \mathcal{L}_{D_{i}}(\mathbf{w}_{i})}{\partial\boldsymbol{\alpha}}\\
&-\eta \nabla_{\boldsymbol{\alpha}}\left(\nabla_{\mathbf{w}_{i}}^{\top}\mathcal{L}_{D^{\prime}_{i}}(\boldsymbol{\alpha}, \mathbf{w}_{i})\nabla_{\mathbf{w}_{i}^{\prime}} \mathcal{L}_{D_{i}}\left(\mathbf{w}_{i}^{\prime}\right)\right).
\end{split}
\label{metagrad}
\end{equation}
The $g_{\mathbf{w}_{i}^{\prime}}\nabla_{\mathbf{w}_{i},\mathbf{w}_{i}}\mathcal{L}_{D^{\prime}_{i}}(\boldsymbol{\alpha}, \mathbf{w}_{i})$ can be calculated as
\begin{equation}
\frac{\nabla_{\mathbf{w}_{i}}\mathcal{L}_{D_{i}^{\prime}}(\boldsymbol{\alpha}, \mathbf{w}_{i}+\epsilon g_{\mathbf{w}_{i}^{\prime}}) - \nabla_{\mathbf{w}_{i}}\mathcal{L}_{D_{i}^{\prime}}(\boldsymbol{\alpha}, \mathbf{w}_{i}-\epsilon g_{\mathbf{w}_{i}^{\prime}})}{2 \epsilon}, 
\label{hessian}
\end{equation}
where $g_{\mathbf{w}_{i}^{\prime}} =\frac{\partial L_{D_i}\left(y, f_{\mathbf{w}_{i, K}\left(\boldsymbol{\alpha}\right)}(\mathbf{X})\right)}{\partial \mathbf{w}_{i,K}\left(\boldsymbol{\alpha}\right)}$ represents the gradient of training loss.

\SetKwInOut{KwIn}{Require}
\begin{algorithm}
\footnotesize{
\caption{\textbf{MMLC} Gradient Computation}\label{outeral}
\KwIn{$P(\mathcal{T})$: distribution over tasks, $\mu, \eta$: learning rate}

Initialize \emph{main model} parameters $\mathbf{w}_{i,0}$ and \emph{meta model} parameters $\boldsymbol{\alpha}$.\\
\While{not done}{
Sample batch of tasks $\mathcal{T}_i \sim P(\mathcal{T})$;\\
\For{all $\mathcal{T}_{i}$}{
\text{Update \emph{main model} parameters $\mathbf{w}_{i}$ by \textbf{Algorithm }\ref{inneral}}\\
}
\text{Update \emph{meta model} parameters $\boldsymbol{\alpha}$ by Eq.~\eqref{metaknowledge}}
}}
\end{algorithm}

\section{\textbf{Experiment}}
\label{experiment}
\subsection{\textbf{Datasets and Setup}}\label{datasets}
\begin{figure}[htbp]
\centerline{\includegraphics[width=0.7\linewidth]{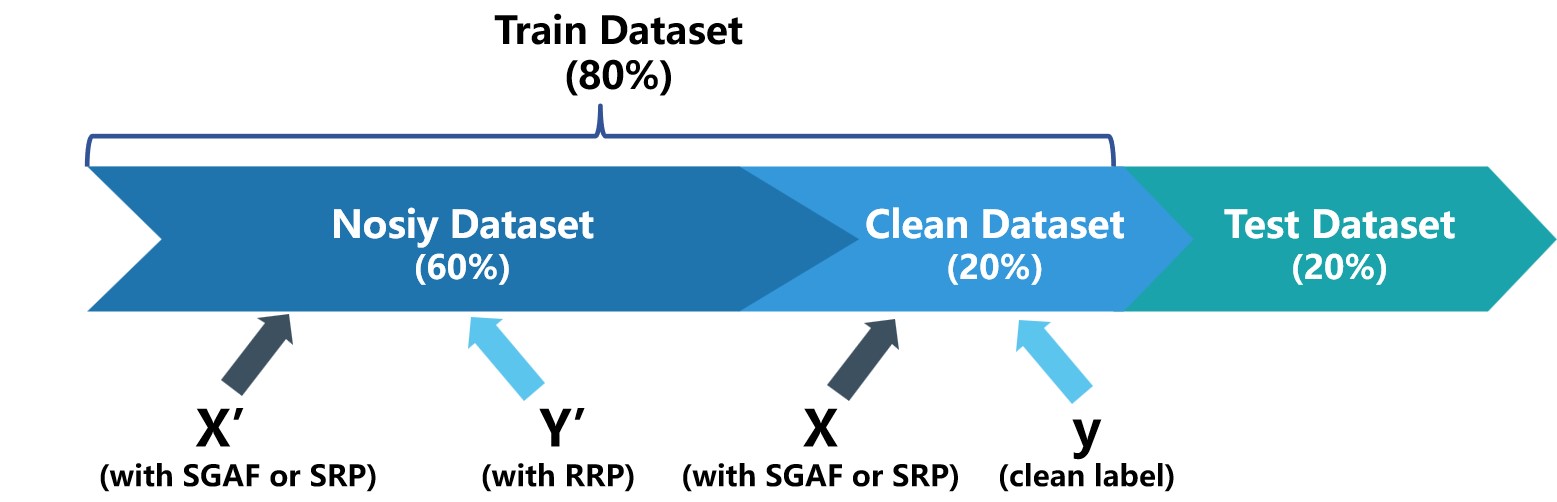}}
\caption{The split structure of dataset for each stock.}
\label{dataset}
\end{figure}

\textbf{Datasets.} Based on recent research \citep{li2022selective, bhandari2022predicting}, we evaluate our method on three different stocks: XOM stock from \textbf{KDD17} \citep{zhang2017stock} public benchmark, American stock index S\&P500 from Yahoo! Finance and Chinese stock index Shangzheng50 (SZ50) from JoinQuant. XOM contains stock from 2007 to 2017, S\&P500 spans from 2010 to 2020, and SZ50 is from 2010 to 2022. The numbers $N$ of XOM,  S\&P500, SZ50 are 2518, 2493, 3159. The specific value of $H$ equals 10, 13 or 15. $n$ is always 30, which means that we apply 30 days for forecasting $H$ days. For the rule of data partitioning in each dataset, the first 60\% time period of the data is used as the noisy data, then the following 40\% time period of data has clean labels, with first half as clean data trained in the meta model, and the other half as testing evaluation(see Fig.~\ref{dataset}). 

During the experiments, we find that the accuracy of forecasting stationary label with the value 1 is the highest. This prompts us to examine the label distribution of clean samples from both clean dataset in meta-training and test dataset. Table \ref{tab3} reveals that the class of label ``1" predominates across all stocks in both datasets. Several publications like \citep{shabani2023augmented} have discussed that stock datasets in the three-class problem will inevitably result in data imbalance, with the largest proportion of stationary class data.

\begin{table*}[ht]
\caption{The distribution of classes of selected stock in the clean dataset of meta training and test dataset on single task of 10-day prediction.$^{\mathrm{a}}$}
\begin{center}
\resizebox{0.8\textwidth}{!}{
\begin{tabular}{*{7}{c}}
  \toprule
  \multirow{2}*{Stock ID } & \multicolumn{3}{c}{Clean Dataset} & \multicolumn{3}{c}{Test Dataset }\\
  \cmidrule(lr){2-4}\cmidrule(lr){5-7}
  &Class 0  &Class 1 &Class 2 &Class 0 &Class 1 &Class 2 \\
  \midrule
  \multirow{1}{*}{XOM} &97 &260 &105 &100 &270 &91  \\
  \multirow{1}{*}{(proportion)} &(20.99\%)   &(56.28\%) &(22.72\%)  &(21.69\%)   &(58.57\%)   &(18.74\%)   \\
  \midrule
  \multirow{1}{*}{S\&P500} &40 & 276 & 141  & 102 & 202 & 152  \\
  \multirow{1}{*}{(proportion)} &(8.75\%)  &(60.39\%) &(30.86\%) &(22.37\%)  &(44.30\%) &(33.33\%)  \\
  \midrule
  \multirow{1}{*}{SZ50} &137 &294 &159  &148 &314 & 127    \\
  \multirow{1}{*}{(proportion)} &(23.22\%)   &(49.83\%)  &(26.95\%)  &(25.13\%) &(53.31\%) & (21.56\%)    \\
  \bottomrule
  \multicolumn{7}{l}{$^{\mathrm{a}}$Class 0: stock price falls. Class 1: stock price remains stable. Class 2: stock price rises.}\\
\end{tabular}}
\label{tab3}
\end{center}
\end{table*}

\textbf{Baseline.} We choose the following state-of-the-art methods for comparison. Here, the settings of \textbf{Resnet32}\footnote{https://github.com/microsoft/MLC} and \textbf{CNN}\footnote{https://github.com/IliaOzhmegov/TradingNeuralNetwork} are the same as the  classifier networks in our proposed model for label correction through normal initialization for weights. The learning rate $\mu, \eta$ are  $3\times10^{-4}$ and $3\times10^{-5}$ with Stochastic Gradient Descent (SGD) optimizer. Moreover, we do experiments for three selected stocks, respectively.

\subsection{\textbf{Experiment results}}\label{results}
\subsubsection{\textbf{Effects of image transformation method}}

We compare our proposed transformation method with traditional methods, employing various models as classifiers in the inner loop. These models are well-known in image classification. Given the limited clean-labeled data, we use identical training and testing datasets with clean labels for all experiments to evaluate accuracy, precision, and F1 score. The hyperparameter settings remain consistent across all experiments.

\begin{table}[ht]
\caption{Performance comparison of \textbf{Resnet32}and \textbf{CNN} on stock dataset}
\begin{center}
\scalebox{0.7}{
\begin{tabular}{*{6}{c}}
\hline
\multirow{2}{*}{Stock ID}&\multicolumn{2}{c}{\multirow{2}{*}{Methods}}& \multicolumn{3}{c}{Error Measures}\\
\cline{4-6}

&\multicolumn{2}{c}{} &
\textbf{\textit{Accuracy}}& \textbf{\textit{Precision}}& \textbf{\textit{F1-score}}  \\
\hline

\multirow{8}{*}{XOM}&\multirow{4}{*}{Resnet32} & GAF & 41.21 & 40.07 &40.57\\
\cline{3-6}

\multirow{8}{*}{}&\multirow{4}{*}{} &RP & 43.82	& 42.72	& 43.16\\
\cline{3-6}

\multirow{8}{*}{}&\multirow{4}{*}{} & SRP &\textbf{48.59}	& \textbf{47.47}	&\textbf{47.40} \\
\cline{3-6}

\multirow{8}{*}{}&\multirow{4}{*}{} &SGAF &47.29	&46.16	& 46.66\\
\cline{2-6}

\multirow{8}{*}{}&\multirow{4}{*}{CNN} &GAF  &43.60	&42.34	&42.85\\
\cline{3-6}

\multirow{8}{*}{}&\multirow{4}{*}{} &RP &43.60	&38.91	&40.84\\
\cline{3-6}

\multirow{8}{*}{}&\multirow{4}{*}{} & SRP   &47.07	&\textbf{43.52}	& 44.91\\
\cline{3-6}

\multirow{8}{*}{}&\multirow{4}{*}{} &SGAF &\textbf{48.81}	&43.49	&\textbf{45.18}\\
\hline

\multirow{8}{*}{S\&P500}&\multirow{4}{*}{Resnet32} &GAF & 39.47 & 33.49  &34.76 \\
\cline{3-6}

\multirow{8}{*}{}&\multirow{4}{*}{} & RP & 42.11 	& 39.71	&38.09 \\
\cline{3-6}

\multirow{8}{*}{}&\multirow{4}{*}{} &SRP   & 44.30	& \textbf{42.73} 	& 41.07 \\
\cline{3-6}

\multirow{8}{*}{}&\multirow{4}{*}{} & SGAF & \textbf{45.83}	& 41.15 & \textbf{41.37}  \\
\cline{2-6}

\multirow{8}{*}{}&\multirow{4}{*}{CNN} & GAF & 39.69 	& 37.79	& 36.14 \\
\cline{3-6}

\multirow{8}{*}{}&\multirow{4}{*}{ } & RP & 37.94 	& 33.81 	& 34.50 \\

\cline{3-6}

\multirow{8}{*}{}&\multirow{4}{*}{ } & SRP   & \textbf{44.29} 	& \textbf{42.00} 	& 39.82  \\
\cline{3-6}

\multirow{8}{*}{}&\multirow{4}{*}{ } & SGAF & 44.08 	& 38.40 	& \textbf{39.87} \\
\hline

\multirow{8}{*}{SZ50}&\multirow{4}{*}{Resnet32} &GAF & 40.58 &41.68  & 40.90\\
\cline{3-6}

\multirow{8}{*}{}&\multirow{4}{*}{} & RP &37.35  & 38.22 & 37.69\\
\cline{3-6}

\multirow{8}{*}{}&\multirow{4}{*}{} & SRP   & 40.91 &40.88  & 40.84 \\
\cline{3-6}

\multirow{8}{*}{}&\multirow{4}{*}{} & SGAF & \textbf{42.78} & \textbf{42.59} & \textbf{42.56}\\
\cline{2-6}

\multirow{8}{*}{}&\multirow{4}{*}{CNN} & GAF &38.03  &37.21 &37.58 \\
\cline{3-6}

\multirow{8}{*}{}&\multirow{4}{*}{ } & RP &41.09  &38.98  &39.81 \\

\cline{3-6}

\multirow{8}{*}{}&\multirow{4}{*}{ } & SRP   &\textbf{43.29}  &40.46  &41.54 \\
\cline{3-6}

\multirow{8}{*}{}&\multirow{4}{*}{ } & SGAF &42.62  &\textbf{41.90}  &\textbf{41.93} \\
\hline
\end{tabular}}
\label{tab1}
\end{center}
\end{table}

From Table \ref{tab1}, we can find that the transformation methods we proposed have better performances for each stock. Nevertheless, even though different optimal visualization methods exist for distinct stock data, they all demonstrate that the trend information in the image is more useful for three-class classification prediction.

\subsubsection{\textbf{ MMLC on Stock movement prediction with single task}}
To better understand the performance of the \textbf{MMLC} model, we conduct a comparative experiment against innovative methods such as \textbf{MLC} \citep{zheng2021meta} and \textbf{IAPTT-GM} \citep{liu2021towards}, as well as a baseline model, \textbf{Resnet32}, which serves as the classifier in the inner loop of both \textbf{MMLC} and \textbf{MLC}. As state-of-the-art methods are typically designed for a single task, we present the results of a single task with a 10-day prediction in this comparative experiment. Additionally, to demonstrate the advantage of correcting noisy labels, the training data for \textbf{Resnet32} only focuses on noisy labels, which differs from data used in Table \ref{tab1}. The results of 13-day prediction and 15-day prediction are shown through Table \ref{appendix13}in \ref{A}.

\begin{table}[ht]
\caption{Test accuracy, precision and F1-score of single task of 10-day prediction on stock dataset}
\begin{center}
\scalebox{0.7}{
\begin{tabular}{*{6}{c}}
\hline
\multirow{2}{*}{Stock ID}&\multicolumn{2}{c}{\multirow{2}{*}{Methods}}& \multicolumn{3}{c}{Error Measures}\\
\cline{4-6}

&\multicolumn{2}{c}{} &
\textbf{\textit{Accuracy}}& \textbf{\textit{Precision}}& \textbf{\textit{F1-score}}  \\
\hline

\multirow{8}{*}{XOM}&\multirow{3}{*}{SRP} & Resnet32 & 33.84 &42.81 &35.86  \\
\cline{3-6}

\multirow{8}{*}{}&\multirow{3}{*}{} & IAPTT-GM &41.86  & 41.55  & 39.47   \\
\cline{3-6}

\multirow{8}{*}{}&\multirow{3}{*}{} & MLC & 52.49  &41.59 & 43.68  \\
\cline{3-6}

\multirow{8}{*}{}&\multirow{3}{*}{} & MMLC &\textbf{57.70}  & \textbf{46.48} & \textbf{43.83}  \\
\cline{2-6}

\multirow{8}{*}{}&\multirow{3}{*}{SGAF} & Resnet32 &36.62 &  40.04 &37.46  \\
\cline{3-6}

\multirow{8}{*}{}&\multirow{3}{*}{} & IAPTT-GM &38.40   & 39.80 &37.10    \\
\cline{3-6}

\multirow{8}{*}{}&\multirow{3}{*}{} & MLC &54.23  &39.92 & 43.61   \\
\cline{3-6}

\multirow{8}{*}{}&\multirow{3}{*}{} & MMLC &\textbf{57.27} & \textbf{41.05} &\textbf{45.28}   \\
\hline
\multirow{8}{*}{S\&P500}&\multirow{3}{*}{SRP} & Resnet32 & 29.82   &32.87 & 30.26  \\
\cline{3-6}

\multirow{8}{*}{}&\multirow{3}{*}{} & IAPTT-GM &36.84   &30.21 &32.09    \\
\cline{3-6}

\multirow{8}{*}{}&\multirow{3}{*}{} & MLC &42.11  &28.34  &30.12    \\
\cline{3-6}

\multirow{8}{*}{}&\multirow{3}{*}{} & MMLC &\textbf{43.86} &\textbf{43.17}  &\textbf{40.30}   \\
\cline{2-6}

\multirow{8}{*}{}&\multirow{3}{*}{SGAF} & Resnet32 &39.69 & 43.72  &40.67  \\
\cline{3-6}

\multirow{8}{*}{}&\multirow{3}{*}{} & IAPTT-GM &42.98 & 44.49  &42.57   \\
\cline{3-6}

\multirow{8}{*}{}&\multirow{3}{*}{} & MLC & 45.18 & 33.24 &35.91  \\
\cline{3-6}

\multirow{8}{*}{}&\multirow{3}{*}{} & MMLC & \textbf{53.07}  & \textbf{49.11} & \textbf{46.96}  \\
\hline
\multirow{8}{*}{SZ50}&\multirow{3}{*}{SRP} & Resnet32 & 35.48 & 41.81   & 37.11 \\
\cline{3-6}

\multirow{8}{*}{}&\multirow{3}{*}{} & IAPTT-GM & 27.84  & 37.45   & 28.48  \\
\cline{3-6}

\multirow{8}{*}{}&\multirow{3}{*}{} & MLC & 46.69  & 38.82  & 40.55    \\
\cline{3-6}

\multirow{8}{*}{}&\multirow{3}{*}{} & MMLC &\textbf{51.10} &\textbf{45.22}  &\textbf{43.97}   \\
\cline{2-6}

\multirow{8}{*}{}&\multirow{3}{*}{SGAF} & Resnet32 & 37.69  & 41.90 &39.10  \\
\cline{3-6}

\multirow{8}{*}{}&\multirow{3}{*}{} & IAPTT-GM & 23.26  & 11.07  & 14.90 \\
\cline{3-6}

\multirow{8}{*}{}&\multirow{3}{*}{} & MLC & 49.41  & 37.27  &39.07 \\
\cline{3-6}

\multirow{8}{*}{}&\multirow{3}{*}{} & MMLC & \textbf{51.78}  & \textbf{42.71} & \textbf{41.18 }  \\
\hline
\end{tabular}}
\label{tab2}
\end{center}
\end{table}

From the results in Table \ref{tab2}, we can observe that as a whole, the MMLC approach we propose has higher accuracy, precision, and F1-score than that of other methods. In particular, the accuracy of the MMLC approach with the SGAF visualization method exceeds 50\% for all three stocks. Furthermore, under both proposed visualization methods, all precision and F1-score metrics exceed 40\%.

\subsubsection{\textbf{ MMLC on Stock movement prediction with multi tasks}}
Multi-task learning in our algorithm can assist us in predicting the stock price for multiple days simultaneously. Hence, we assign tasks with 10-day, 13-day, and 15-day predictions to achieve long-term predictions. Additionally, since our method is model-agnostic, we can employ a variety of current classification models in the inner loop. In our experiments, we employ \textbf{Resnet32} and \textbf{CNN} as the classifier networks. The experimental results are shown in Tables \ref{tab4} and \ref{tab5}.

\begin{table}[htbp]
\caption{Test accuracy, precision and F1-score of multiple task on stock dataset with \textbf{Resnet32}}
\begin{center}
\scalebox{0.7}{
\begin{tabular}{*{6}{c}}
\hline
\multirow{2}{*}{Stock ID}&\multicolumn{2}{c}{\multirow{2}{*}{Methods}}& \multicolumn{3}{c}{Error Measures}\\
\cline{4-6}

&\multicolumn{2}{c}{} &
\textbf{\textit{Accuracy}}& \textbf{\textit{Precision}}& \textbf{\textit{F1-score}}  \\
\hline

\multirow{6}{*}{XOM}&\multirow{3}{*}{SRP+MMLC} & \textbf{10-day} &57.80    &49.06   &44.99     \\
\cline{3-6}

\multirow{6}{*}{}&\multirow{3}{*}{} & \textbf{13-day} &60.44    &\textbf{51.29}    &46.59      \\
\cline{3-6}

\multirow{6}{*}{}&\multirow{3}{*}{} & \textbf{15-day} &\textbf{61.76 }   &45.43    &\textbf{47.71}      \\
\cline{2-6}

\multirow{6}{*}{}&\multirow{3}{*}{SGAF+MMLC} &\textbf{10-day}  &58.46    &\textbf{49.28 }   &46.36    \\
\cline{3-6}

\multirow{6}{*}{}&\multirow{3}{*}{} &\textbf{13-day} &59.34    &46.88    &46.34    \\
\cline{3-6}

\multirow{6}{*}{}&\multirow{3}{*}{} & \textbf{15-day}  &\textbf{60.66 }   &40.65   &\textbf{47.29 }  \\
\hline
\multirow{6}{*}{S\&P500}&\multirow{3}{*}{SRP+MMLC} & \textbf{10-day}  &44.44    &44.11    &37.05 \\
\cline{3-6}

\multirow{6}{*}{}&\multirow{3}{*}{} & \textbf{13-day}  &\textbf{75.33}   &\textbf{56.88}    &\textbf{64.82}    \\
\cline{3-6}

\multirow{6}{*}{}&\multirow{3}{*}{} & \textbf{15-day}  &74.22   &55.09 &63.24      \\
\cline{2-6}

\multirow{6}{*}{}&\multirow{3}{*}{SGAF+MMLC} &\textbf{10-day}  &47.11    &43.98    &41.11    \\
\cline{3-6}

\multirow{6}{*}{}&\multirow{3}{*}{} &\textbf{13-day} &71.78    &\textbf{60.06}    &\textbf{64.33 }   \\
\cline{3-6}

\multirow{6}{*}{}&\multirow{3}{*}{} & \textbf{15-day}  &\textbf{72.00 }   &59.33    &63.49   \\
\hline
\multirow{6}{*}{SZ50}&\multirow{3}{*}{SRP+MMLC} & \textbf{10-day}  &52.49     &43.24    &41.00    \\
\cline{3-6}

\multirow{6}{*}{}&\multirow{3}{*}{} & \textbf{13-day} &52.83     &40.85    &42.30    \\
\cline{3-6}

\multirow{6}{*}{}&\multirow{3}{*}{} & \textbf{15-day}  &\textbf{54.72}     & \textbf{45.70}  &\textbf{45.15}       \\
\cline{2-6}

\multirow{6}{*}{}&\multirow{3}{*}{SGAF+MMLC} & \textbf{10-day} &51.46     &44.64    &43.09     \\
\cline{3-6}

\multirow{6}{*}{}&\multirow{3}{*}{} &\textbf{13-day}  &50.09     &42.04    &43.72        \\
\cline{3-6}

\multirow{6}{*}{}&\multirow{3}{*}{} & \textbf{15-day}  &\textbf{52.66}     &\textbf{45.12}    &\textbf{46.31}      \\
\hline
\end{tabular}}
\label{tab4}
\end{center}
\end{table}

Compared to Table \ref{tab2}, which shows 10-day single-task predictions using the Resnet32 classifier, our algorithm can also perform effectively for 10-day forecasting with the multi-task loss. For instance, with the SGAF transformation method, the accuracy for XOM stock price, S\&P500 stock price, and SZ50 stock price under multi-task learning can reach 58.46\%, 47.11\%, and 51.46\%, respectively,  versus 57.27\%, 53.07\%, and 51.78\% in single-task learning. Moreover, nearly all precision and F1 scores for the selected stocks exceed 41\% and some of them might even be higher than the result of single-task learning.

\begin{table}[htbp]
\caption{Test accuracy, precision and F1-score of multiple task on stock dataset with \textbf{CNN}}
\begin{center}
\scalebox{0.7}{
\begin{tabular}{*{6}{c}}
\hline
\multirow{2}{*}{Stock ID}&\multicolumn{2}{c}{\multirow{2}{*}{Methods}}& \multicolumn{3}{c}{Error Measures}\\
\cline{4-6}

&\multicolumn{2}{c}{} &
\textbf{\textit{Accuracy}}& \textbf{\textit{Precision}}& \textbf{\textit{F1-score}}  \\
\hline

\multirow{6}{*}{XOM}&\multirow{3}{*}{SRP+MMLC} & \textbf{10-day} &55.60    &41.95   &47.19    \\
\cline{3-6}

\multirow{6}{*}{}&\multirow{3}{*}{} & \textbf{13-day} &56.26    &42.98   &48.08   \\
\cline{3-6}

\multirow{6}{*}{}&\multirow{3}{*}{} & \textbf{15-day} &\textbf{57.58}    &\textbf{43.82}   &\textbf{49.15}    \\
\cline{2-6}

\multirow{6}{*}{}&\multirow{3}{*}{SGAF+MMLC} &\textbf{10-day}  &\textbf{57.80}     &\textbf{54.99}   &46.57   \\
\cline{3-6}

\multirow{6}{*}{}&\multirow{3}{*}{} &\textbf{13-day} &57.14    &54.18   &48.48    \\
\cline{3-6}

\multirow{6}{*}{}&\multirow{3}{*}{} & \textbf{15-day}  &56.70     &54.61    &\textbf{51.13}  \\
\hline
\multirow{6}{*}{S\&P500}&\multirow{3}{*}{SRP+MMLC} & \textbf{10-day}  &45.11    &43.71    &37.60  \\
\cline{3-6}

\multirow{6}{*}{}&\multirow{3}{*}{} & \textbf{13-day}  &\textbf{73.78 }   & 61.67    &\textbf{65.66 }    \\
\cline{3-6}

\multirow{6}{*}{}&\multirow{3}{*}{} & \textbf{15-day}  &\textbf{73.78}   &\textbf{63.69}    &64.91      \\
\cline{2-6}

\multirow{6}{*}{}&\multirow{3}{*}{SGAF+MMLC} &\textbf{10-day}  &52.22     &40.06    &45.21     \\
\cline{3-6}

\multirow{6}{*}{}&\multirow{3}{*}{} &\textbf{13-day} &\textbf{75.78}     &\textbf{75.45 }   &\textbf{65.77}    \\
\cline{3-6}

\multirow{6}{*}{}&\multirow{3}{*}{} & \textbf{15-day} &73.56     &60.71    &63.68    \\
\hline
\multirow{6}{*}{SZ50}&\multirow{3}{*}{SRP+MMLC} & \textbf{10-day}  &52.49     &\textbf{46.89}    &40.01   \\
\cline{3-6}

\multirow{6}{*}{}&\multirow{3}{*}{} & \textbf{13-day}  &53.17    &43.35    &42.29     \\
\cline{3-6}

\multirow{6}{*}{}&\multirow{3}{*}{} & \textbf{15-day}  &\textbf{54.72}     &45.31    &\textbf{44.57}       \\
\cline{2-6}

\multirow{6}{*}{}&\multirow{3}{*}{SGAF+MMLC} & \textbf{10-day} & 51.97    &45.37    &42.13     \\
\cline{3-6}

\multirow{6}{*}{}&\multirow{3}{*}{} &\textbf{13-day} &\textbf{53.34}     &\textbf{48.44}    &\textbf{45.70}       \\
\cline{3-6}

\multirow{6}{*}{}&\multirow{3}{*}{} & \textbf{15-day}  &52.66     &45.25    &44.39     \\
\hline
\end{tabular}}
\label{tab5}
\end{center}
\end{table}

Table \ref{tab5} presents the prediction results obtained with the CNN classifier, demonstrating the model-agnostic nature of our algorithm. For example, for XOM stock, when using the SRP transformation method and the Resnet32 classifier, we achieve accuracy of 57.80\%, 60.44\%, and 61.76\% for 10-day, 13-day, and 15-day predictions, respectively. When replacing Resnet32 with CNN, we can achieve accuracy of 55.60\%, 56.26\%, and 57.58\% for XOM stock, respectively. For 10-day, 13-day, and 15-day F1-scores of XOM stock with SRP transformation method, Resnet can separately reach 44.99\%, 46.59\% and 47.71\%, and CNN can separately get 47.19\%, 48.08\% and 49.15\% as well. Hence, any classification model can also be suitable.

\section{Conclusion}
\label{conclusion}

We investigate a novel research issue for financial time series, with the aim of effectively correcting labels through adaptive labeling. For the data processing, we propose two visualization methods for historical data called SGAF and SRP without the tendency confusion. Moreover, we also learn the stock's intrinsic patterns (denoted as $\mathbf{Y}^{\prime}$ in the noisy data) in the prediction horizon to achieve better adaptive labeling. We establish the label corrector model with two branches to extract effective features. With multi-task learning, we can evaluate the classification performance of each task, while obeying the same rule of labeling trending direction from the pattern ``Y". With this shared meta knowledge, there is no need to retrain the label correction model for each task. Through various comparative experiments, we demonstrate the superior performance and the model-agnostic nature of our \textbf{MMLC} model. Nevertheless, several open problems are still left. For instance, we encounter the challenge of imbalanced datasets, which could be solved by an adaptive loss function in the future. Another challenge is that the initialization of weights and hyper-parameters will cause different results sometimes. We plan to optimize the initial parameters in the next project. Code is available on GitHub \footnote{https://github.com/senyuanya/MMLC}.

\section*{Acknowledgment}
We would like to thank Yubin Lu and Yufu Lan for the helpful discussions. This work was partly done at the Dongguan Key Laboratory for Data Science and intelligent Medicine. This work was supported by National Key Research and Development Program of China 2021ZD0201300, National Natural Science Foundation of China (NSFC) 12141107, Fundamental Research Funds for the Central Universities 5003011053 and Fundamental Research Funds for the Central Universities, HUST: 2023JYCXJJ045.
\appendix
\setcounter{table}{0}
\section{}
\label{A}
\begin{algorithm}
\caption{Generation of images ``X" and ``Y"}\label{geimage}
\SetAlgoLined
\footnotesize{
\KwIn{Time series data for each stock $x_{1},x_{2}, \cdots, x_{N}$ with data size $N$, window size $n$, the length of prediction horizon $H$.}
\KwOut{Image set of ``X" and ``Y". }
\For{$k=0$ to $N-n-(H-1)$}{
\text{Reprocessing by SGAF and SRP:}\\
\quad\quad Calculate $G_{k}$ in Eq.~\eqref{GM} and visualize $\widetilde{G_{k}}$ in Eq.~\eqref{mgaf}\\
\quad\quad Calculate $R_{k}$ in Eq.~\eqref{rpmatrix} and visualize $\widetilde{R_{k}}$ in Eq.~\eqref{mrp} \\
\text{Reprocessing by RRP:}\\
\quad\quad Visualize $\{ratio_{k}(1),\cdots,ratio_{k}(H)\}$ through Eq.~\eqref{RR}\\
}
}
\end{algorithm}

\begin{table}[ht]
\caption{ Test accuracy, precision and F1-score of single task of 13-day (15-day) prediction on stock dataset}
\begin{center}
\scalebox{0.7}{
\begin{tabular}{*{6}{c}}
\hline
\multirow{2}{*}{Stock ID}&\multicolumn{2}{c}{\multirow{2}{*}{Methods}}& \multicolumn{3}{c}{Error Measures}\\
\cline{4-6}

&\multicolumn{2}{c}{} &
\textbf{\textit{Accuracy}}& \textbf{\textit{Precision}}& \textbf{\textit{F1-score}}  \\
\hline

\multirow{8}{*}{XOM}&\multirow{3}{*}{SRP} & Resnet32 &31.66 (31.21) &42.48 (45.29)  &34.16 (34.67)  \\
\cline{3-6}

\multirow{8}{*}{}&\multirow{3}{*}{} & IAPTT-GM &37.99 (40.00)  &37.63 (44.18)  &37.03 (39.60)   \\
\cline{3-6}

\multirow{8}{*}{}&\multirow{3}{*}{} & MLC &53.49 (60.44)  &40.88 (44.41)  &44.96 (48.13)    \\
\cline{3-6}

\multirow{8}{*}{}&\multirow{3}{*}{} & MMLC &\textbf{54.15} (\textbf{62.42})	  &\textbf{46.20 } (\textbf{48.86}) &\textbf{48.74} (\textbf{48.82})   \\
\cline{2-6}

\multirow{8}{*}{}&\multirow{3}{*}{SGAF} & Resnet32 &36.46 (31.65)  &47.38 (44.54) &39.04 (34.80)   \\
\cline{3-6}

\multirow{8}{*}{}&\multirow{3}{*}{} & IAPTT-GM &40.18 (40.66) &37.92 (49.70) &38.48 (42.52)       \\
\cline{3-6}

\multirow{8}{*}{}&\multirow{3}{*}{} & MLC &57.64 (56.48)	 &45.46 (47.01)  &47.61 (49.34)     \\
\cline{3-6}

\multirow{8}{*}{}&\multirow{3}{*}{} & MMLC & \textbf{60.48} (\textbf{60.66}) &\textbf{54.35} (\textbf{53.57})  &\textbf{53.08}  (\textbf{52.82})    \\
\hline
\multirow{8}{*}{S\&P500}&\multirow{3}{*}{SRP} & Resnet32 &40.39 (42.44)  &69.09 (69.93)  &45.76 (48.05)    \\
\cline{3-6}

\multirow{8}{*}{}&\multirow{3}{*}{} & IAPTT-GM &45.03 (43.33) &53.94 (50.94)  &48.48 (46.14)    \\
\cline{3-6}

\multirow{8}{*}{}&\multirow{3}{*}{} & MLC &\textbf{75.50} (74.22) &57.00 (55.09) &64.96  (63.24) \\
\cline{3-6}

\multirow{8}{*}{}&\multirow{3}{*}{} & MMLC &75.06 (\textbf{74.44}) &\textbf{65.29} (\textbf{65.11})  &\textbf{66.84 } (\textbf{63.84})     \\
\cline{2-6}

\multirow{8}{*}{}&\multirow{3}{*}{SGAF} & Resnet32 &40.84 (39.78) &67.24 (66.00) &45.53 (45.12) \\
\cline{3-6}

\multirow{8}{*}{}&\multirow{3}{*}{} & IAPTT-GM &45.03 (46.22)  &53.94  (52.99)  &48.48 (58.77)  \\
\cline{3-6}

\multirow{8}{*}{}&\multirow{3}{*}{} & MLC &75.50 (74.44) &66.33 (68.33) & 65.74 (64.15)   \\
\cline{3-6}

\multirow{8}{*}{}&\multirow{3}{*}{} & MMLC &\textbf{75.94} (\textbf{74.67  }) &\textbf{68.46} (\textbf{68.54}) &\textbf{66.42}  (\textbf{65.01})   \\
\hline
\multirow{8}{*}{SZ50}&\multirow{3}{*}{SRP} & Resnet32 &31.45 (31.90) &40.12 (41.17) &33.33  (33.51)  \\
\cline{3-6}

\multirow{8}{*}{}&\multirow{3}{*}{} & IAPTT-GM &53.68 (38.42) &35.32 (35.63) &40.62 (36.01)   \\
\cline{3-6}

\multirow{8}{*}{}&\multirow{3}{*}{} & MLC &50.77 (36.88) &40.87 (41.43) &41.32 (38.14)     \\
\cline{3-6}

\multirow{8}{*}{}&\multirow{3}{*}{} & MMLC &\textbf{52.48} (\textbf{47.86}) &\textbf{46.67} (\textbf{44.84}) &\textbf{47.09} (\textbf{45.97})    \\
\cline{2-6}

\multirow{8}{*}{}&\multirow{3}{*}{SGAF} & Resnet32 &33.68 (30.01) &41.47 (40.26) &35.86 (32.17)\\
\cline{3-6}

\multirow{8}{*}{}&\multirow{3}{*}{} & IAPTT-GM &49.40 (41.00) &36.67 (37.95 ) &41.75  (38.69)  \\
\cline{3-6}

\multirow{8}{*}{}&\multirow{3}{*}{} & MLC &50.77 (39.62) & 38.24 (42.51) &40.54 (40.57)   \\
\cline{3-6}

\multirow{8}{*}{}&\multirow{3}{*}{} & MMLC &\textbf{51.97} (\textbf{47.17}) &\textbf{41.02} (\textbf{47.42}) & \textbf{42.22} (\textbf{47.24} ) \\
\hline
\end{tabular}}
\label{appendix13}
\end{center}
\end{table}

\bibliographystyle{elsarticle-num-names} 
\bibliography{references}

\end{document}